\documentclass{article}

\PassOptionsToPackage{numbers, sort&compress}{natbib}
\usepackage[main,final]{neurips_2026}

\usepackage{amsmath, amssymb, amsfonts}
\usepackage{booktabs}
\usepackage{graphicx}
\usepackage{hyperref}
\usepackage{url}
\usepackage{xcolor}
\usepackage{caption}
\usepackage{multirow}
\usepackage{microtype}
\usepackage{tikz}
\usetikzlibrary{positioning, arrows.meta, shapes.geometric, calc, fit, backgrounds, decorations.pathmorphing}
\usepackage{pgfplots}
\pgfplotsset{compat=1.18}
\usepgfplotslibrary{fillbetween}

\hypersetup{
    colorlinks=true,
    linkcolor=blue!60!black,
    citecolor=green!50!black,
    urlcolor=blue!60!black,
}

\title{How Much Is One Recurrence Worth? \\ Iso-Depth Scaling Laws for Looped Language Models}

\author{%
  Kristian Schwethelm\textsuperscript{1} \And
  Daniel R\"uckert\textsuperscript{1,2,3} \And
  Georgios Kaissis\textsuperscript{4} \AND
  {\normalfont \textsuperscript{1}Chair for AI in Healthcare and Medicine, Technical University of Munich (TUM) and} \\ {TUM University Hospital, Munich, Germany} \\
  {\normalfont \textsuperscript{2}Department of Computing, Imperial College London, UK} \\
  {\normalfont \textsuperscript{3}Munich Center for Machine Learning (MCML), Germany} \\
  {\normalfont \textsuperscript{4}Hasso Plattner Institute for Digital Engineering, University of Potsdam, Germany} \\
}

\begin{document}
\maketitle
\begin{abstract}
We measure how much one recurrence is worth to a looped transformer in equivalent unique parameters. From an iso-depth pretraining sweep across recurrence counts $r \in \{1, 2, 4, 8\}$ spanning ${\sim}50\times$ in training compute, we fit a joint scaling law $L = E + A\,(N_\text{once} + r^{\varphi} N_\text{rec})^{-\alpha} + B\,D^{-\beta}$ and measure a \emph{recurrence-equivalence exponent} $\varphi {=} 0.46$. Intuitively, $\varphi$ tells us whether looping a block $r$ times is equivalent in validation loss to $r$ unique blocks of a non-looped model (full equivalence, $\varphi{=}1$) or to a single block run repeatedly with no capacity gain ($\varphi{=}0$). Our $\varphi {=} 0.46$ sits in between, so replacing unique blocks with shared recurrences \emph{increases} validation loss at matched training compute. For example, at $r{=}4$ a 410M looped model performs on par with a 580M non-looped model, but incurs the training cost of a 1B non-looped one. We demonstrate the utility of $\varphi$ as a diagnostic tool on two case studies: commonly used truncated backpropagation lowers $\varphi$ to $0.38$, indicating that the loop mechanism is poorly trained under truncation, even though validation loss decreases. Conversely, hyperconnections raise $\varphi$ to $0.65$, a genuine capacity gain. Our method separates true loop improvements from training-side gains, a distinction raw validation loss cannot make. Code is available at \url{https://github.com/kschwethelm/looped-lm-scaling}.
\end{abstract}

\section{Introduction}
\label{sec:intro}

\emph{Can a transformer block looped $r$ times replace $r$ non-looped blocks at matched compute budget?} Looped transformers iterate a shared block of layers, increasing model depth without adding new layers (parameters)~\citep{dehghani2019universal,fan2024looped}. Such models have recently drawn attention in language modelling as a route to implicit, latent-space reasoning and test-time compute scaling~\citep{geiping2025scaling, zhu2025scaling, bae2025mor, fu2025thinkathard, mcleish2025retrofitted, koishekenov2025etd, prairie2026parcae}. However, in practice, most looped LMs use only small recurrence counts, revealing a cost: applying a block $r$ times costs the same compute (FLOPs) as applying $r$ unique blocks (i.e., a non-looped model of the same depth). A looped model therefore performs the same number of layer computations with fewer unique parameters. To what extent repeated applications of shared parameters can recover the capacity provided by unique blocks remains unclear. We measure this using an iso-depth scaling analysis.

For this, we compare model performance across recurrence counts $r \in \{1, 2, 4, 8\}$, where $r{=}1$ is a conventional non-looped transformer. As $r$ increases, we reduce the number of layers in the shared recurrent block so that every architecture performs the same number of layer computations per token, making them \emph{iso-depth} (Figure~\ref{fig:architecture-overview}). At matched model width, the compared models then have approximately the same per-token training and inference FLOPs, while their number of unique parameters decreases by a factor of up to $3.2$ (Figure~\ref{fig:teaser}, left). We evaluate this trade-off with a pretraining scaling law across six compute budgets from $4.64 {\times} 10^{17}$ to $2.15 {\times} 10^{19}$~FLOPs.

The standard Chinchilla scaling law $L(N,D)$~\citep{hoffmann2022training} can relate validation loss to unique parameters $N$ and training tokens $D$, but has limited utility in comparing looped models. The recurrence count $r$ does not enter the law, and the parameter count $N$ has different FLOPs cost under looping. We therefore introduce a joint scaling law $L(N_\text{once}, N_\text{rec}, D, r)$ that includes $r$ and splits $N$ into the shared recurrent parameters ($N_\text{rec}$) and the single-use ones ($N_\text{once}$), with $N_\text{once} + N_\text{rec} = N$.

At the core of our joint scaling law, we introduce a \emph{recurrence-equivalence exponent} $\varphi$ that quantifies how much each shared parameter contributes to loss reduction relative to a unique one: $L(N_\text{once}, N_\text{rec}, D, r) = E + A\,(N_\text{once} + r^{\varphi} N_\text{rec})^{-\alpha} + B\,D^{-\beta}$. $\varphi$ has two natural reference points. $\varphi = 1$ attributes full parameter equivalence to each recurrence (no sharing cost), so at matched training FLOPs the looped model would reach the same validation loss as the non-looped baseline. $\varphi = 0$ indicates no gain through recurrences (a pure sharing cost). For our baseline architecture, we find $\varphi = 0.46$ (Figure~\ref{fig:teaser}). This means a 410M $r{=}4$ model performs on par with a 580M non-looped model, but incurs the training cost of a 1B non-looped one (derivation in Appendix~\ref{app:equiv-sizes}).

\emph{Is $\varphi$ fixed?} No. We propose the shift $\Delta\varphi$ as a diagnostic tool and demonstrate it on two case studies: truncated backpropagation (gradients are only computed for the last few loops)~\citep{geiping2025scaling,mcleish2025retrofitted, prairie2026parcae} and hyperconnections (parallel residual streams between loops)~\citep{zhu2025hyperconnections}. Both decrease validation loss but move $\varphi$ in opposite directions. Truncated backpropagation decreases $\varphi$ to $0.38$, so each loop becomes \emph{less} powerful. Hyperconnections raise $\varphi$ to $0.65$. We further show that $\Delta\varphi$ distinguishes whether a validation-loss gain comes from a better loop mechanism or a hidden trade-off with inference cost.

\begin{figure}[!t]
\centering
\input{figs/teaser}
\caption{\emph{Left:} at matched effective depth, per-token forward FLOPs $F(r)$ stay flat while unique parameters $N(r)$ drop as $r$ grows. Shared parameter capacity $N_\text{once} + r^{\varphi} N_\text{rec}$ with $\varphi{=}0.46$ drops more slowly. \emph{Right:} compute-optimal validation loss per architecture against compute budget $C$. Empirical per-budget optima (crosses) track our $\varphi{=}0.46$ fit (solid curves). The standard form ($\varphi{=}1$) collapses all four architectures onto a single curve (dashed), which fits none of the empirical results.}
\label{fig:teaser}
\end{figure}

\paragraph{Contributions.}
\begin{enumerate}
    \item We introduce a \emph{recurrence-equivalence exponent} $\varphi$ that quantifies, via a joint scaling law over $r$, how many unique parameter blocks one recurrence is worth, and measure a baseline of $\varphi = 0.46$: each recurrence is worth roughly half a unique block at the same FLOPs.
  \item We propose $\Delta\varphi$ as a diagnostic tool for comparing looped LM training recipes and architectures. It separates architecture-side from training-side gains, with the compute-optimal allocation reflecting the resulting deployment cost.
  \item By measuring $\Delta\varphi$, we show that commonly used truncated backpropagation weakens looping and raises inference cost ($\varphi=0.38$), while hyperconnections provide a genuine capacity gain that also narrows the compute-optimum ($\varphi=0.65$).
\end{enumerate}

\section{Related Work}
\label{sec:related}

\paragraph{Looped language models.}
Huginn~\citep{geiping2025scaling} and Ouro~\citep{zhu2025scaling} have scaled looped LMs to ${\sim}3$B parameters and trillion-token training budgets with strong downstream results, often matching much larger dense transformers. However, this comes at a proportional training-compute cost, as each loop multiplies forward-backward FLOPs. Compute-matched comparisons to non-looped models at single training budgets report a consistent pattern: looped models trail non-looped baselines on validation loss and parametric-knowledge tasks but close the gap or outperform them on reasoning benchmarks~\citep{saunshi2025understanding}. We extend these findings to the scaling-law setting with various training budgets. Unlike per-architecture scaling laws~\citep{kaplan2020scaling,hoffmann2022training}, our joint law fits looped and non-looped models together under a single recurrence-equivalence exponent $\varphi$. Architectural and training-efficiency methods such as retrofitting~\citep{mcleish2025retrofitted,koishekenov2025etd}, adaptive compute~\citep{bae2025mor, fu2025thinkathard, knupp2026dreamer, jeddi2026loopformer}, truncated backpropagation~\citep{geiping2025scaling, prairie2026parcae, mcleish2025retrofitted}, and hyperconnections~\citep{zhu2025hyperconnections, zeitounHyperloopTransformers2026} affect $\varphi$. We apply our framework to truncated backpropagation and hyperconnections, leaving the others to future work. See Appendix~\ref{app:extended_related} for extended discussion.

\paragraph{Looped scaling laws.}
Concurrent work~\citep{prairie2026parcae} fits a scaling law at fixed parameter count to identify compute optimal recurrence, motivated by keeping weight memory overhead fixed. However, in this setting, per-token inference FLOPs and KV cache memory grow with recurrence compared to the baseline, which can dominate inference cost (Appendix~\ref{app:inference-cost}). In contrast, our iso-depth setup keeps both costs approximately fixed and characterises compute-optimal behaviour under this constraint. We additionally focus on directly measuring the per-recurrence parameter-sharing cost.

\section{Methodology}
\label{sec:setup}

\begin{figure}[t]
  \centering
  \colorlet{archmuted}{black!55}
  \colorlet{archframe}{black!30}
  \colorlet{archpre}{gray!22}
  \colorlet{archrecur}{orange!38}
  \colorlet{archinject}{green!30}
  \colorlet{archstate}{yellow!48}
  \colorlet{archbase}{blue!20}
  \colorlet{archloop}{red!65!black}
  \resizebox{\linewidth}{!}{%
  \begin{tikzpicture}[
    x=1cm,
    y=1cm,
    module/.style={draw=archframe, line width=0.45pt, rounded corners=1.4pt, minimum height=0.56cm, minimum width=0.76cm, inner sep=1.5pt, align=center, font=\scriptsize},
    prelude/.style={module, fill=archpre},
    recurrent/.style={module, fill=archrecur},
    injection/.style={module, fill=archinject, minimum width=0.60cm},
    coda/.style={module, fill=archpre},
    state/.style={module, fill=archstate, minimum height=0.30cm, minimum width=0.82cm, font=\tiny},
    baseline/.style={module, fill=archbase, minimum width=1.10cm},
    token/.style={draw=archframe, line width=0.4pt, rounded corners=1.2pt, fill=gray!8, inner xsep=3pt, inner ysep=2pt, font=\scriptsize\ttfamily},
    paneltitle/.style={font=\small\bfseries, anchor=west},
    header/.style={font=\fontsize{6.3pt}{6.3pt}\selectfont, text=black!80, align=center},
    smalllabel/.style={font=\tiny, text=archmuted},
    collabel/.style={font=\scriptsize\bfseries, align=center},
    flow/.style={draw=archmuted, line width=0.45pt, -{Stealth[length=2.5pt, width=2.2pt]}},
    feedback/.style={draw=archmuted, line width=0.4pt, densely dashed, -{Stealth[length=2.5pt, width=2.2pt]}},
    loopflow/.style={draw=archloop, line width=0.55pt, -{Stealth[length=2.7pt, width=2.3pt]}}
  ]
    \node[paneltitle] at (-0.65,5) {Looped next-token prediction};
    \node[header] at (1.12,4.25) {Prelude};
    \node[header] at (3.18,4.25) {Shared recurrent block};
    \node[header] at (5.12,4.25) {Coda};

    \newcommand{\predictionrow}[2]{%
      \node[prelude] (pre#1) at (1.12,#2) {$P$};%
      \node[recurrent, minimum width=0.58cm] (rec#1a) at (2.05,#2) {$1$};%
      \node[recurrent, minimum width=0.58cm] (rec#1b) at (2.78,#2) {$2$};%
      \node[smalllabel, text=black!70] (dots#1) at (3.50,#2) {$\cdots$};%
      \node[recurrent, minimum width=0.58cm] (rec#1r) at (4.16,#2) {$r$};%
      \node[coda] (cod#1) at (5.12,#2) {$C$};%
      \draw[flow] (pre#1.east) -- (rec#1a.west);%
      \draw[flow] (rec#1a.east) -- (rec#1b.west);%
      \draw[flow] (rec#1b.east) -- (dots#1.west);%
      \draw[flow] (dots#1.east) -- (rec#1r.west);%
      \draw[flow] (rec#1r.east) -- (cod#1.west);%
    }

    \def\rowone{3.48}
    \def\rowtwo{2.18}
    \def\rowthree{0.88}

    \node[token, align=center, anchor=east] (prompt) at (0.63,\rowone) {What is\\12$\times$5?};
    \node[smalllabel, anchor=south] at ($(prompt.north)+(0,0.04)$) {context};
    \predictionrow{1}{\rowone}
    \draw[flow] (prompt.east) -- (pre1.west);
    \draw[flow] (cod1.east) -- (5.95,\rowone);
    \node[smalllabel, anchor=south] at (5.72,{\rowone+0.08}) {$x_1$};
    \node[token, anchor=west] (tok1) at (5.95,\rowone) {It};

    \predictionrow{2}{\rowtwo}
    \draw[flow] (cod2.east) -- (5.95,\rowtwo);
    \node[smalllabel, anchor=south] at (5.72,{\rowtwo+0.08}) {$x_2$};
    \node[token, anchor=west] (tok2) at (5.95,\rowtwo) {is};

    \predictionrow{3}{\rowthree}
    \draw[flow] (cod3.east) -- (5.95,\rowthree);
    \node[smalllabel, anchor=south] at (5.72,{\rowthree+0.08}) {$x_3$};
    \node[token, anchor=west] (tok3) at (5.95,\rowthree) {60};

    \draw[feedback, rounded corners=2pt]
      (tok1.east) -- (6.82,\rowone) -- (6.82,{\rowtwo+0.57}) -- (1.12,{\rowtwo+0.57}) -- (pre2.north);
    \draw[feedback, rounded corners=2pt]
      (tok2.east) -- (6.82,\rowtwo) -- (6.82,{\rowthree+0.57}) -- (1.12,{\rowthree+0.57}) -- (pre3.north);
    \node[smalllabel, fill=white, inner sep=1pt] at (4.10,{\rowtwo+0.57}) {sampled token};

    \node[paneltitle] at (7.35,5) {Iso-depth architecture family};

    \def\unit{0.175}
    \def\xone{8.02}
    \def\xtwo{10.02}
    \def\xfour{12.02}
    \def\xeight{14.02}
    \def\stackbase{0.43}

    \node[baseline, minimum height={20*\unit cm}] (base1) at (\xone,{\stackbase+10*\unit}) {20 layers\\unshared\\(no loop)};
    \node[collabel, anchor=south] at (\xone,4.12) {$r{=}1$};

    \node[prelude, minimum width=1.10cm, minimum height={2*\unit cm}] (p2) at (\xtwo,{\stackbase+1*\unit}) {2 prelude};
    \node[injection, minimum width=1.10cm, minimum height=0.25cm, above=1pt of p2] (i2) {inject};
    \node[recurrent, minimum width=1.10cm, minimum height={8*\unit cm}, above=1pt of i2] (r2) {8 recur\\layers};
    \node[state, minimum width=1.10cm, above=1pt of r2] (h2) {$h^{(t)}$};
    \node[coda, minimum width=1.10cm, minimum height={2*\unit cm}, above=1pt of h2] (c2) {2 coda};
    \node[collabel, anchor=south] at (\xtwo,4.12) {$r{=}2$};
    \draw[loopflow, rounded corners=2pt] ($(h2.east)+(0.03,0)$) -| ($(i2.east)+(0.30,0)$) -- ($(i2.east)+(0.03,0)$);
    \node[smalllabel, text=archloop, anchor=west] at ($(r2.east)+(0.32,0)$) {$\times2$};

    \node[prelude, minimum width=1.10cm, minimum height={2*\unit cm}] (p4) at (\xfour,{\stackbase+1*\unit}) {2 prelude};
    \node[injection, minimum width=1.10cm, minimum height=0.25cm, above=1pt of p4] (i4) {inject};
    \node[recurrent, minimum width=1.10cm, minimum height={4*\unit cm}, above=1pt of i4] (r4) {4 recur\\layers};
    \node[state, minimum width=1.10cm, above=1pt of r4] (h4) {$h^{(t)}$};
    \node[coda, minimum width=1.10cm, minimum height={2*\unit cm}, above=1pt of h4] (c4) {2 coda};
    \node[collabel, anchor=south] at (\xfour,4.12) {$r{=}4$};
    \draw[loopflow, rounded corners=2pt] ($(h4.east)+(0.03,0)$) -| ($(i4.east)+(0.30,0)$) -- ($(i4.east)+(0.03,0)$);
    \node[smalllabel, text=archloop, anchor=west] at ($(r4.east)+(0.32,0)$) {$\times4$};

    \node[prelude, minimum width=1.10cm, minimum height={2*\unit cm}] (p8) at (\xeight,{\stackbase+1*\unit}) {2 prelude};
    \node[injection, minimum width=1.10cm, minimum height=0.25cm, above=1pt of p8] (i8) {inject};
    \node[recurrent, minimum width=1.10cm, minimum height={2*\unit cm}, above=1pt of i8] (r8) {2 recur};
    \node[state, minimum width=1.10cm, above=1pt of r8] (h8) {$h^{(t)}$};
    \node[coda, minimum width=1.10cm, minimum height={2*\unit cm}, above=1pt of h8] (c8) {2 coda};
    \node[collabel, anchor=south] at (\xeight,4.12) {$r{=}8$};
    \draw[loopflow, rounded corners=2pt] ($(h8.east)+(0.03,0)$) -| ($(i8.east)+(0.30,0)$) -- ($(i8.east)+(0.03,0)$);
    \node[smalllabel, text=archloop, anchor=west] at ($(r8.east)+(0.32,0)$) {$\times8$};
  \end{tikzpicture}%
  }
  \caption{\emph{Left:} each next-token prediction $x_i$ runs the prelude once, the shared recurrent block $r$ times, and the coda once. Input injection between recurrences is omitted for clarity.
  \emph{Right:} our architectures keep the effective depth fixed, e.g., to 20 by reducing the recurrent block from 8 to 4 to 2 layers as $r$ increases from 2 to 4 to 8. The $r{=}1$ is a conventional 20-layer non-looped transformer.}
  \label{fig:architecture-overview}
\end{figure}

\subsection{Looped Transformer Architecture}
\label{sec:specification}

We compare four transformer variants: a non-looped baseline ($r{=}1$) and looped models with $r \in \{2, 4, 8\}$ recurrences. All looped variants follow the prelude-recur-coda architecture~\citep{geiping2025scaling}. Here, an unshared prelude first embeds the tokens, then the shared recurrent block refines the hidden state $h^{(t)}$ with $r$ loops, and finally the unshared coda predicts the next token (Figure~\ref{fig:architecture-overview}). This allows some layers to specialize in input processing and prediction while parameter sharing is confined to the middle of the network. We fix the effective depth at $\ell_\text{eff} = \ell_\text{prelude} + r \cdot \ell_\text{recur} + \ell_\text{coda} = 20$. Setting $(\ell_\text{prelude}, \ell_\text{coda}) = (2, 2)$ gives a recurrent block of $\ell_\text{recur} = 16/r$ layers, corresponding to $(8, 4, 2)$ layers for $r \in \{2, 4, 8\}$. Width is parameterised as $d_\text{model} = 64s$ for an integer scale factor $s$, with attention head dimension 128. Following~\citet{geiping2025scaling}, a linear input-injection layer combines the prelude output with the recurrent state before each recurrence. Its small FLOPs overhead is included in every iso-FLOPs comparison. Full architectural details are provided in Appendix~\ref{app:architecture}.

Let $n_b {=} 12 d^2$ be the parameter count of a single transformer layer at width $d {\equiv}d_\text{model}$ (four $d {\times} d$ attention projections plus the $d \to 4d \to d$ MLP), and $n_i {=} 2 d^2$ that of the injection layer. We write $N$ for the transformer's total non-embedding parameters and adopt the standard convention that per-token forward FLOPs ${\approx} 2N$ and training FLOPs ${\approx}6N$~\citep{kaplan2020scaling}. Our reported FLOPs also include parameter-free attention matmuls ($12\,h q T$ at training, with $h$ heads of dimension $q$ and sequence length $T$) and the unembedding matmul ($6 d V$ at training, with vocabulary $V$), both fixed across architectures at matched width. For simplicity we exclude them from the equations below.

For looped models, we split $N$ into parameters that are used once, $N_\text{once} = (\ell_\text{prelude} + \ell_\text{coda})\,n_b$, and those that are recurring, $N_\text{rec} = \ell_\text{recur}\,n_b + n_i$ (fully-looped architectures correspond to $N_\text{once} = 0$). In our setup $\ell_\text{recur} = 16/r$, so each additional loop shrinks the recurrent block. Thus, our looped models with $r \in \{2, 4, 8\}$ have ${\sim}61\%$, ${\sim}41\%$, and ${\sim}31\%$ of the parameters of a same-width non-looped model.\footnote{$N(r) = (4 + 16/r)\,n_b + n_i = (48 + 192/r + 2)\,d^2$, giving $\{146, 98, 74\}\,d^2$ for $r \in \{2,4,8\}$ versus $240\,d^2$ at $r{=}1$.} For example, at $s{=}10$, $N \in \{98.3, 59.8, 40.2, 30.3\}$~M.

In terms of compute, the looped transformer executes the prelude once, the recurrent block $r$ times (each preceded by the injection layer), and the coda once, so per-token forward FLOPs are:
\begin{equation}
  F_\text{fwd}(r) = 2(N_\text{once} + r\,N_\text{rec}) \approx F_\text{fwd}(1) = 2\,\ell_\text{eff}\,n_b = 2N.
  \label{eq:forward-flops}
\end{equation}
Thus, all variants use the same $F_\text{fwd}$ up to a small injection layer overhead of $r/120 \in \{1.7\%, 3.3\%, 6.7\%\}$ at $r \in \{2, 4, 8\}$.\footnote{The overhead is $2 r n_i / (2 \ell_\text{eff} n_b) = r \cdot 2d^2 / (20 \cdot 12 d^2) = r/120$.} Training FLOPs are $F_\text{train}(r) = 3\,F_\text{fwd}(r)$, so at a fixed compute budget $C$ every variant trains on approximately the same token count $D \approx C / F_\text{train}$.

\subsection{Joint Scaling Law}
\label{sec:joint-law-def}

The standard Chinchilla scaling law is defined as
\begin{equation}
    L(N, D) = E + A N^{-\alpha} + B D^{-\beta},
    \label{eq:chinchilla}
\end{equation}
where $L$ is validation loss (nats), $N$ the non-embedding parameter count, $D$ training tokens, $E$ the irreducible loss, and $A, B, \alpha, \beta$ fitted constants. The law assumes loss decreases with higher $N$ and $D$. The two terms split the loss into a parameter contribution $A N^{-\alpha}$ and a token contribution $B D^{-\beta}$, with $\alpha, \beta$ setting how fast each diminishes and (via their ratio) the compute-optimal allocation between $N$ and $D$~\citep{hoffmann2022training}. Our iso-compute design holds $D$ approximately fixed across $r$ but reduces $N$, so the same token count is spread over fewer parameters as $r$ grows.

To isolate how much the looped parameters $N_\text{rec}$ effectively contribute per recurrence, we extend the Chinchilla form (Equation~\ref{eq:chinchilla}) with a \emph{recurrence-equivalence exponent} $\varphi \geq 0$ acting on $N_\text{rec}$:
\begin{equation}
    L(N_\text{once}, N_\text{rec}, D, r) = E + A \, \bigl(N_\text{once} + r^{\varphi} N_\text{rec}\bigr)^{-\alpha} + B \, D^{-\beta}.
    \label{eq:joint-3d}
\end{equation}
We refer to $N_\text{eff} = N_\text{once} + r^{\varphi} N_\text{rec}$ as the effective parameter count. The factor $r^\varphi \geq 1$ amplifies the contribution of the looped parameters to loss reduction since they are used multiple times. 

Intuitively, $N_\text{eff}$ is the parameter count of a non-looped model that would match the looped variant's loss at the same $D$. So, if looped models outperform non-looped models at same $N$, we attribute the performance gain to looped parameters contributing more than they would in a single forward pass. Note that the law is fit on iso-compute runs, so $C$ enters only implicitly through the data. The non-looped reference matches loss at the same $D$, not at the same compute. When $\varphi < 1$, the recurrent block's contribution to capacity grows slower with $r$ than its contribution to compute (Equation~\ref{eq:forward-flops}), so looped variants underperform non-looped models in compute-matched comparisons.

$\varphi$ has two natural reference points. $\varphi = 0$ means the recurrent block contributes the same to loss whether run once or $r$ times. The extra $r{-}1$ executions add FLOPs without any capacity gain, and increasing $r$ at fixed compute only shrinks $N$ and thus raises loss. $\varphi = 1$ corresponds to the iso-FLOPs non-looped model, where each recurrence contributes as much as equivalent unique parameter blocks, so all four architectures would perform the same. Values $0 < \varphi < 1$ quantify partial recovery. $\varphi > 1$ would require the looped block to contribute more than unique blocks (eg, $r{=}8$ outperforms $r{=}4$), which we do not observe.

\section{Scaling Law Results}
\label{sec:scaling}

\subsection{Experimental Details}
\label{sec:experimental-details}

We train at six compute budgets and various model sizes $s {=} d_\text{model}/64$ and adjust training tokens $D$ accordingly. Although our iso-depth setup only requires architectures at the same size to have the same effective depth, we hold effective depth fixed at 20 layers throughout this sweep. Appendix~\ref{app:depth-allocation} tests 18- and 24-layer configurations. Jointly scaling width and depth introduces additional choices over the layer split between the prelude, recurrent block, and coda, which we discuss in Section~\ref{sec:discussion}. At each compute budget, the token count is determined by $D = C/(3F_{\mathrm{fwd}})$, so wider models train on fewer tokens (Table~\ref{tab:grid}). On the resulting 116 pretraining runs, we first fit per-architecture Chinchilla laws, then our proposed joint law with $\varphi$.

\paragraph{Implementation.} Our implementation builds on nanochat~\citep{karpathy2025nanochat}: a decoder-only transformer with RMSNorm~\citep{zhang2019rms}, RoPE~\citep{su2023roformer}, QK normalisation~\citep{dehghani2023scaling}, and squared-ReLU MLPs~\citep{so2021primer}. We use FlashAttention-2 \& 3~\citep{dao2024flashattention2, shah2024flashattention3} as the attention backends.

\paragraph{Optimisation.} Matrix parameters are optimised with MuonH~\citep{wen2025hyperball, jordan2024muon}. Embedding, unembedding, and norm parameters are optimised with AdamW~\citep{loshchilov2019adamw}. Weight decay is set to zero (first-order no-op under MuonH's Frobenius-sphere constraint~\citep{ren2026muonh}). Learning rates transfer across width, batch size, and training horizon via muP~\citep{yang2021mup} and HyperP~\citep{ren2026muonh}. In Appendix~\ref{app:hparams}, we sweep base LR and batch size at the reference width and find optima agreeing across architectures (LR regret below $0.005$~nats per architecture). All runs use base MuonH LR $\eta_\text{base} = 0.014$ and AdamW base LRs $0.3$ (embedding), $0.004$ (unembedding), and $0.005$ (norm), with each LR linearly decayed to $10\%$ of its peak.

\paragraph{Data and validation.} Training data is a subset of FineWeb-Edu~\citep{lozhkov2024fineweb-edu}, tokenised with the Llama~2 tokenizer~\citep{touvron2023llama2} ($32{,}000$ base vocabulary) and pre-packed into fixed-length sequences of $2{,}049$ tokens (the extra token provides the next-token target for position $2{,}048$). Thus, all four architectures see exactly the same data stream. Validation loss is reported in nats on a held-out FineWeb-Edu split packed identically to training.

\subsection{Per-Architecture Chinchilla Fits}
\label{sec:scaling-fits}

\begin{figure}[tb]
    \centering
    \includegraphics[width=\textwidth]{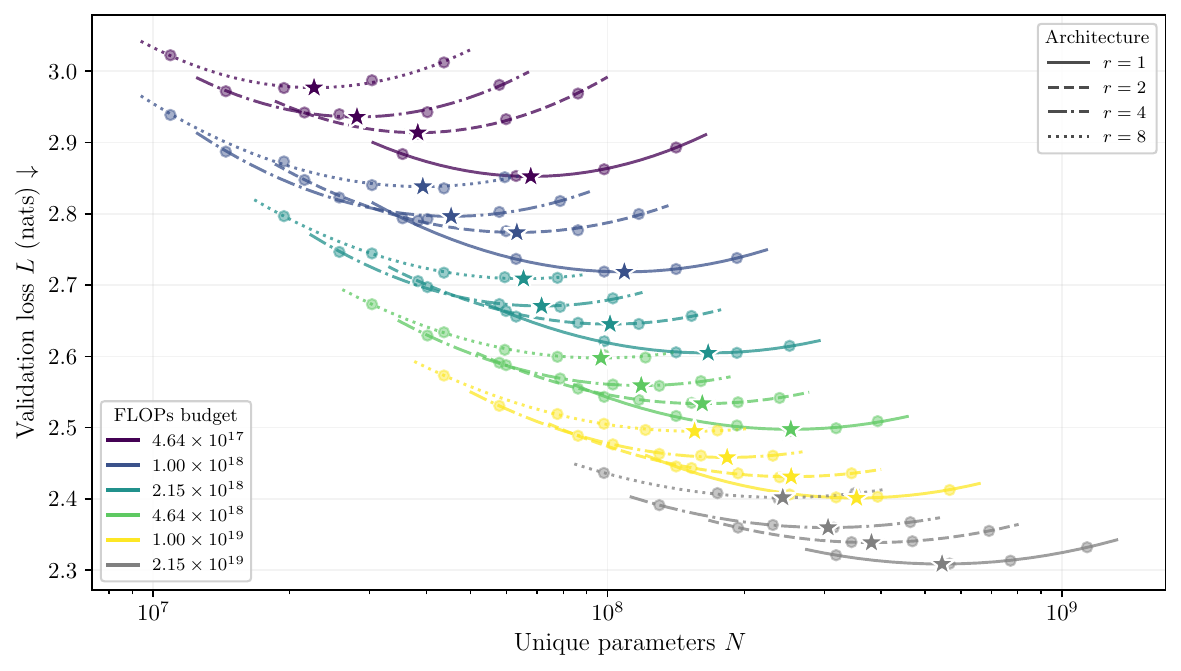}
    \caption{Scaling curves at fixed compute budgets. Curves are per-(budget, $r$) parabolic fits in $\log N$. Stars mark the fitted compute-optimal $(N^*, L^*)$ points. See Figure~\ref{fig:isoflops-width} for an alternative view.}
    \label{fig:isoflops}
\end{figure}

Figure~\ref{fig:isoflops} shows validation loss against unique non-embedding parameters $N$ at our grid. At fixed compute $C$ all four architectures trace an approximately parabolic iso-FLOPs curve in $\log N$. The parabolas are systematically offset upward for larger $r$, with looped minima at wider $s$ than the non-looped baseline (Appendix Figure~\ref{fig:isoflops-width}). We fit the standard Chinchilla scaling law (Equation~\ref{eq:chinchilla}) separately for each architecture to characterise individual scaling behaviour and limitations.

\paragraph{Fitting protocol.}
We follow~\citet{hoffmann2022training} and minimise the Huber loss~\citep{huber1964robust} ($\delta = 10^{-3}$) between predicted and empirical \emph{log} validation loss using L-BFGS~\citep{nocedal1980lbfgs}. Specifically, we parameterise the law in log-space ($a = \log A$, $b = \log B$, $e = \log E$) and minimise
\begin{equation}
    \mathcal{L}(a,\alpha,b,\beta,e) = \sum_i \mathrm{Huber}_{\delta}\!\left(\mathrm{LSE}\bigl(a - \alpha \log N_i,\; b - \beta \log D_i,\; e\bigr) - \log L_i\right),
    \label{eq:huber-obj}
\end{equation}
where $\mathrm{LSE}$ is log-sum-exp. The log-space objective treats relative errors uniformly across the wide dynamic range of $N$ and $D$. Because the objective is non-convex, we take the best of 500 random L-BFGS-B restarts (details in Appendix~\ref{app:fit-residuals}). Table~\ref{tab:fit-params} reports the fitted parameters per architecture, all achieving $R^2 > 0.997$ (predicted-vs-actual scatter and residuals in Appendix~\ref{app:fit-residuals}).

\begin{table}[tb]
\centering
\caption{Chinchilla scaling-law fit parameters per architecture. Amplitudes $A, B$ are rounded to 2 significant figures as they are only loosely identified under iso-Compute~\citep{besiroglu2024chinchilla}.}
\label{tab:fit-params}
\small
\begin{tabular}{@{}lrrrrrrr@{}}
\toprule
Arch. & $A$ & $\alpha$ & $B$ & $\beta$ & $E$ & Huber\,$(\times 10^{-5})$ & $R^2$ \\
\midrule
$r{=}1$ & $58$  & $0.251$ & $150$  & $0.267$ & $1.56$ & 5.84 & 0.9979 \\
$r{=}2$ & $33$  & $0.216$ & $910$  & $0.365$ & $1.60$ & 5.27 & 0.9983 \\
$r{=}4$ & $23$  & $0.191$ & $1300$ & $0.388$ & $1.56$ & 6.81 & 0.9976 \\
$r{=}8$ & $41$  & $0.235$ & $780$  & $0.362$ & $1.69$ & 5.33 & 0.9980 \\
\bottomrule
\end{tabular}
\end{table}

\begin{figure}[tb]
    \centering
    \includegraphics[width=\textwidth]{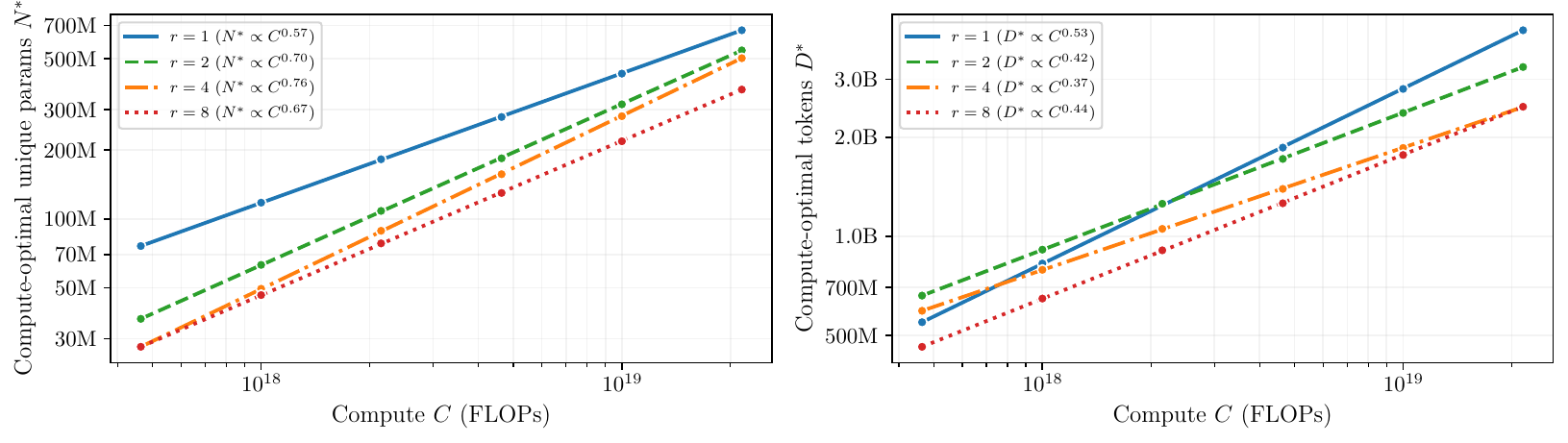}
    \caption{Compute-optimal allocation per architecture. Left: optimal unique parameter count $N^*(C)$ with fitted exponent in the legend. Right: optimal training tokens $D^*(C)$.}
    \label{fig:optimal-alloc}
\end{figure}

\paragraph{Compute-optimal allocation and gap.}
The fitted exponents in Table~\ref{tab:fit-params} characterise each architecture's scaling behaviour in isolation but are not directly comparable across $r$. The unique parameter count $N$ has different semantics under parameter sharing (each parameter in the recurrent block contributes $r$ times for looped models), and iso-FLOPs sampling places each architecture in a different region of the $(N, D)$ plane, where $\alpha$ and $\beta$ are weakly identified~\citep{besiroglu2024chinchilla}. The compute-optimal loss frontier $L^*_r(C)$, by contrast, is directly comparable across $r$. We derive the optimal parameter and token allocation $N^*(C), D^*(C)$ by minimising $L(N, C / F(N))$ over $N$ at each budget, with $F(N)$ the architecture's empirical per-token FLOPs at $N$. The results are shown in Figure~\ref{fig:optimal-alloc}. Looped optima favour wider models than the non-looped baseline, but their unique parameter count is still lower (left). The optimum compensates for parameter sharing by widening, which raises per-token FLOPs and therefore lowers the optimal training-token count (right).

The resulting loss frontier trails the baseline by $[0.03, 0.06]$~nats at $r{=}2$, $[0.05, 0.08]$~nats at $r{=}4$, and $[0.09, 0.12]$~nats at $r{=}8$, growing monotonically with $r$ across the six budgets. The gap widens at lower budgets and flattens at our largest ($\Delta \leq 0.006$~nats between $10^{19}$ and $2.15 \times 10^{19}$~FLOPs).

\paragraph{Extrapolation beyond the grid.}
To test whether the gap persists past our grid, we train $r{=}1$ and $r{=}4$ models at $s{=}34$ on $47$B tokens (${\sim}4{\times} 10^{20}$~FLOPs, ${\sim}20\times$ the top of our grid). The looped model trails by $0.061$~nats in validation loss, inside the $[0.05, 0.08]$~nats $r{=}4$ band measured across the grid. See Appendix \ref{app:extrapolation} for detailed results.

\subsection{Joint Scaling Law Fit $\varphi$}
\label{sec:joint-3d}

We now fit the joint law of Equation~\ref{eq:joint-3d} across all 116 runs. The recurrence-equivalence exponent $\varphi$ places every architecture on a common $(N_\text{once} + r^\varphi N_\text{rec}, D)$ surface and measures how much each recurrence amplifies the contribution of $N_\text{rec}$. This also reduces the four per-architecture fits above (20 parameters) to one shared law with 6 parameters $(A, \alpha, B, \beta, E, \varphi)$. We minimise the same Huber-on-log objective as Equation~\ref{eq:huber-obj}. The results are shown in Table~\ref{tab:joint-3d}.

\begin{table}[tb]
\centering
\caption{Joint $(N_\text{once}, N_\text{rec}, D, r)$ scaling law (Equation~\ref{eq:joint-3d}) fit. The free-$\varphi$ row reports 95\% block-bootstrap CIs (200 resamples of (budget, architecture) cells) below the $\varphi$ point estimate. Amplitudes $A, B$ are only loosely identified under iso-compute designs~\citep{besiroglu2024chinchilla} and are omitted from the table.}
\label{tab:joint-3d}
\small
\setlength{\tabcolsep}{5pt}
\begin{tabular}{@{}lccccc@{}}
\toprule
Form & $\alpha$ & $\beta$ & $E$ & $\varphi$ & $R^2$ \\
\midrule
Joint, $\varphi$ free &
  $0.199$ &
  $0.369$ &
  $1.57$ &
  \shortstack{$0.459$\\ \scriptsize$[0.41, 0.53]$} &
  $0.9972$ \\[0.6em]
Restricted ($\varphi = 0$, pure sharing cost)              & $0.227$ & $0.390$ & $1.71$ & $0.00$ & $0.9858$ \\
Restricted ($\varphi = 1$, non-looped equivalence)     & $0.218$ & $0.410$ & $1.66$ & $1.00$ & $0.9552$ \\
\bottomrule
\end{tabular}
\end{table}

We measure $\varphi = 0.46$, well below the non-looped reference of $\varphi = 1$ but clearly above $0$. Each recurrence amplifies the looped parameters' contribution to loss reduction by a factor of $r^{0.46}$, so the recurrent block contributes ${\sim}1.9\times$ its unique parameter count at $r{=}4$ and ${\sim}2.6\times$ at $r{=}8$. This matches the qualitative pattern from the compute-optimal allocation above. Each recurrence increases the capacity of the looped parameters ($\varphi > 0$), requiring fewer tokens to reach equal performance at same $N$. That is why the optimum can spend a smaller share of its FLOPs on tokens and more on widening the model. However, the added capacity is only partial ($\varphi < 1$): the $r^{0.46}$ amplification of the looped block's contribution does not keep up with the linear $r$ increase in its FLOPs cost. Per-FLOP capacity therefore falls with $r$, and the looped frontier trails the non-looped baseline.

\paragraph{Robustness.} The $95\%$ block-bootstrap CI on $\varphi$ is $[0.41, 0.53]$, with no resample reaching $\varphi = 0$ or $\varphi = 1$. The $R^2$ values in Table~\ref{tab:joint-3d} look uniformly high because most variance across the runs comes from the compute-budget axis (cross-architecture loss spans only ${\sim}0.1$~nats). On that scale, small drops from $0.997$ (free $\varphi$) are substantial. Figure~\ref{fig:teaser} (right) visualises this fit-quality gap for $\varphi=1$. Per-$r$ residuals (Appendix Table~\ref{tab:joint-residuals-per-r}) are comparable across architectures (RMSE $0.009$--$0.011$~nats). Refits on the low-budget half ($C \leq 2.15 \times 10^{18}$) and high-budget half ($C \geq 4.64 \times 10^{18}$) give $\varphi = 0.44$ and $\varphi = 0.49$, both inside the CI. See the full analysis in Appendix~\ref{app:fit-residuals}.
\section{Case Studies}
\label{sec:case_studies}

The fitted $\varphi {=} 0.46$ describes our baseline recipe and architecture. We also measure $\varphi$ under two candidate interventions: truncated backpropagation (a training-recipe change) and hyperconnections (an architectural change). For each case study we rerun the iso-FLOPs grid for $r \in \{2, 4, 8\}$ at our four lower budgets ($4.64 {\times} 10^{17}$ to $4.64 {\times} 10^{18}$~FLOPs), reuse the unchanged $r{=}1$ runs as the baseline, and refit the joint law (Table~\ref{tab:probe_fits}). The resulting $\Delta\varphi$ is a single-number summary of how much each recurrence gains or loses capacity under the method. Implementation details are in Appendix~\ref{app:probing}.

\subsection{Truncated Backpropagation}
\label{sec:case_truncbptt}

Truncated backpropagation through time (BPTT) is a training-efficiency method that is commonly applied to looped transformers~\citep{geiping2025scaling,mcleish2025retrofitted, prairie2026parcae}. Under full BPTT, gradients flow backward through all $r$ recurrences. Truncated BPTT detaches the recurrent state for all but the last $r_\text{bwd}$ loops, so the early recurrences run forward only and skip the backward pass. We follow~\citet{prairie2026parcae} and set $r_\text{bwd} = \lceil r/2 \rceil$. In our setup, the skipped backward passes save roughly $30\%$ of the per-token training FLOPs, allowing more training tokens at fixed budget.

The scaling curves in Figure~\ref{fig:trunc_hc_isoflops} (left) show that truncated BPTT substantially lowers validation loss across all runs, suggesting at first glance that the method works well. However, the measured $\varphi$ disagrees, dropping from $0.45$ to $0.38$ (Table~\ref{tab:probe_fits}). So each additional recurrence is now worth much less in unique-parameter terms than before (smaller $r^\varphi$ in $N_\text{eff}$, so matching the same $N_\text{eff}$ now requires a larger $N_\text{rec}$). This likely reflects early loops no longer receiving an accurate learning signal. Evidence for this is the $r{=}2$ architecture, which consistently has the largest residual error (Figure~\ref{fig:probing_fit_quality}). Here $r_\text{bwd}{=}1$, so the shared block receives a direct gradient only from the second recurrence. The first recurrence still runs forward and conditions the second loop's input, but this indirect signal is evidently too weak to train the looping mechanism. The same applies to the forward-only loops at $r{=}4$ and $r{=}8$, which lowers the per-recurrence utility and pulls $\varphi$ down. Refitting without $r{=}2$ leaves $\varphi$ essentially unchanged at $0.37$ (Table~\ref{tab:probe_fits}), confirming that the capacity drop is not an $r{=}2$ artifact. Overall, the joint law attributes the validation-loss improvement to a larger token budget and a wider compute-optimal model that offsets the per-loop capacity loss but raises inference cost.

\subsection{Hyperconnections}
\label{sec:case_hyperconnections}

We replace the linear input injection of our baseline model with hyperconnections~\citep{zhu2025hyperconnections}, which scale and mix $K$ parallel residual lanes across loops. We hypothesize that better information flow inside the recurrent block leads to a better loop mechanism. We use $K{=}2$ lanes and full BPTT.

\begin{figure}[tb]
    \centering
    \includegraphics[width=\textwidth]{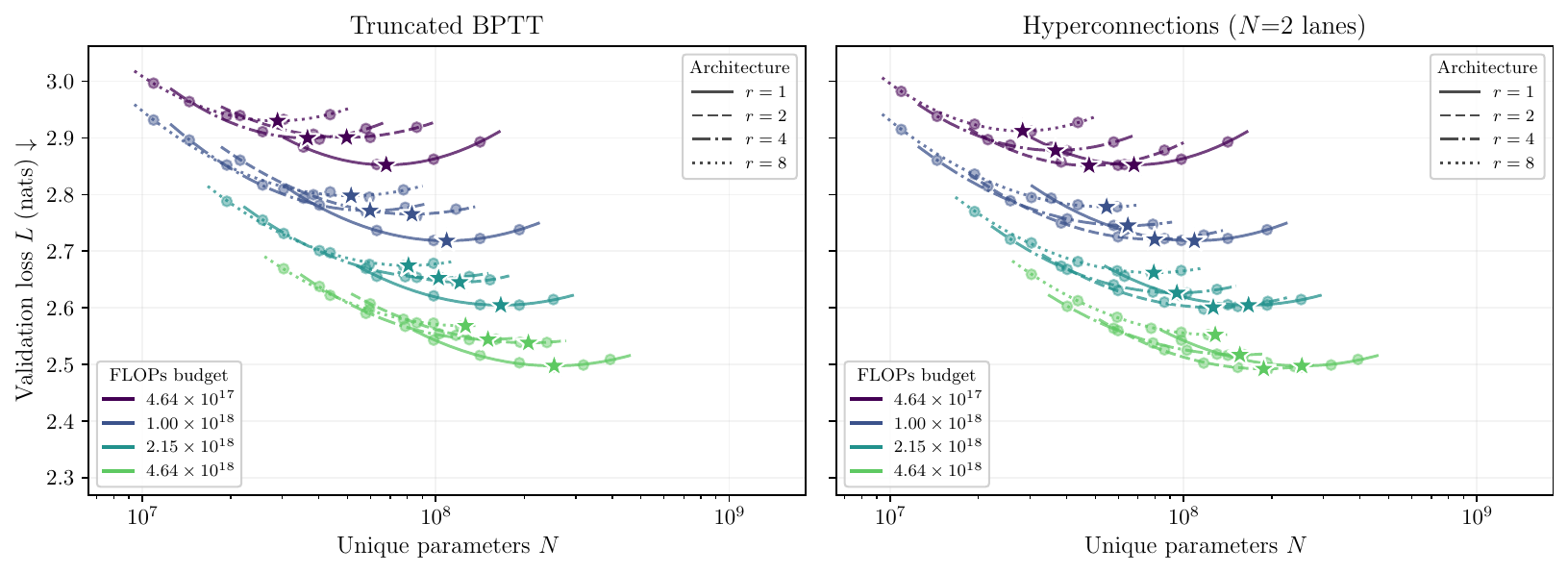}
    \caption{Scaling curves under the two case studies. Thin curves are per-(budget, $r$) parabolic fits in $\log N$. Stars mark the fitted compute-optimal $(N^*, L^*)$ points.}
    \label{fig:trunc_hc_isoflops}
\end{figure}

The scaling curves in Figure~\ref{fig:trunc_hc_isoflops} (right) show that hyperconnections substantially lower validation loss across all looped runs, and $\varphi$ jumps from $0.45$ to $0.65$ (Table~\ref{tab:probe_fits}). Each additional recurrence is now worth much more in unique-parameter terms than before. Interestingly, the $r{=}2$ architecture even matches or beats the $r{=}1$ baseline at some budgets. Note that $\varphi{=}1$ would require all four architectures to lie on the same compute-optimal frontier, not just $r{=}2$ beating $r{=}1$ at a few budgets.

Thus, hyperconnections are a genuine architectural improvement, as validation loss falls and the compute-optimal allocation moves to narrower widths, lowering per-token inference FLOPs (the opposite of the widening seen under truncated BPTT). However, hyperconnections could in principle be applied to the $r{=}1$ baseline as well, replacing the residual connections between transformer blocks~\citep{zhu2025hyperconnections}. Our case study applies them only at the loop boundary, which is the natural site for a within-loop intervention. An $r{=}1$ baseline modified the same way would likely shift the calibration of $\varphi$. However, our primary comparisons are between looped models, with the non-looped baseline serving only as a calibration anchor. The shift would also likely be modest: concurrent work~\citep{zeitounHyperloopTransformers2026} shows loop-level hyperconnections beat layer-level hyperconnections.

\begin{table}[tb]
\centering
\caption{Joint-law refits for each case study alongside the full-BPTT baseline. The baseline was refitted to the four lower budgets. The truncated-BPTT (excl.\ $r{=}2$) row excludes the $r{=}2$ model, which is an outlier because its shared block receives a direct gradient from only one recurrence.}
\label{tab:probe_fits}
\small
\begin{tabular}{@{}lccccc@{}}
\toprule
Variant & $\alpha$ & $\beta$ & $E$ & $\varphi$ & $R^2$ \\
\midrule
Baseline (full BPTT, linear injection)               & $0.266$ & $0.457$ & $1.85$ & $0.453$ & $0.9959$ \\
Truncated BPTT                                       & $0.255$ & $0.484$ & $1.87$ & $0.380$ & $0.9827$ \\
Truncated BPTT (excl.\ $r{=}2$)                      & $0.265$ & $0.492$ & $1.89$ & $0.373$ & $0.9958$ \\
Hyperconnections                          & $0.308$ & $0.464$ & $1.93$ & $0.646$ & $0.9900$ \\
\bottomrule
\end{tabular}
\end{table}

\section{Discussion}
\label{sec:discussion}

\paragraph{Per-recurrence value.}
Our $\varphi{=}0.46$ quantifies how much one recurrence is worth in equivalent unique parameters. At $r{=}4$ the shared block recovers $4^{0.46} {\approx} 1.89$ unique blocks of capacity, about $47\%$ of full equivalence ($\varphi{=}1$). We asked whether a block looped $r$ times can replace $r$ non-looped blocks at matched compute. Our measurement says no, not under our recipe. But $\varphi$ is not a property of looping in general. It reflects the joint state of architecture, optimiser, and training recipe, and the case studies of Section~\ref{sec:case_studies} show $\varphi$ can move substantially under either type of change.

\paragraph{$\Delta\varphi$ as a development metric.}
$\Delta\varphi$ separates token-side from architecture-side gains. Our two case studies illustrate the two cases. Truncated BPTT lowers loss on the token channel but drops $\varphi$, while hyperconnections lower loss on the capacity channel by raising $\varphi$. We therefore recommend $\Delta\varphi$ as a diagnostic tool alongside validation loss for any new looped LM recipe or architecture. Measuring it is relatively cheap. Per architecture, a focused case study of ${\sim}20$ runs across our four lower budgets totals ${\sim}5 {\times} 10^{19}$~FLOPs. The same runs also yield the per-architecture compute-optimal allocation. Other methods worth quantifying include shrinking the shared fraction (larger prelude/coda), per-token adaptive compute~\citep{bae2025mor, fu2025thinkathard}, retrofitting pretrained non-looped models~\citep{mcleish2025retrofitted, koishekenov2025etd}, and training with a diffusion objective in place of unrolling the loops~\citep{shingDiffusionBlocksBlockwiseNeural2026}.

\paragraph{Inference cost.}
$\varphi$ also predicts inference cost. A higher $\varphi$ means each loop adds more capacity, so the same loss can be reached with fewer unique parameters. The freed compute then goes into more training tokens, giving a narrower compute-optimum and lower per-token inference FLOPs. A lower $\varphi$ has the opposite effect. Each loop adds less, so the optimum widens to compensate and trains on fewer tokens, raising inference cost. A method that substantially lowers $\varphi$ should therefore be treated cautiously, since the same validation loss is reachable more cheaply with fewer, more powerful recurrences. This is exactly what we observe for truncated BPTT (Section~\ref{sec:case_truncbptt}). 

\paragraph{Scaling width and depth.}
Our iso-depth setup fixes the effective depth across recurrence counts at a given model size, so the $r{=}\{1,2,4,8\}$ architectures have approximately the same per-token compute (Appendix \ref{app:inference-cost}). It does not constrain the depth used at different model sizes, so an iso-depth sweep can scale width and depth together as long as the depth remains matched across $r$ at each size. Scaling the number of layers in a looped model is complicated by how those layers are split between the prelude, recurrent block, and coda. A layer inside the recurrent block is executed $r$ times, while prelude and coda layers are executed once, so layer count and effective depth scale differently depending on where a layer is added. In our primary sweep, we vary width and hold effective depth fixed at 20 layers to isolate parameter sharing from these architectural choices. Appendix~\ref{app:depth-allocation} tests model scaling with 18 and 24 effective layers and yields a similar $\varphi{=}0.47$, supporting our results.

\paragraph{Downstream validation.}
Applying our scaling law to downstream continuation loss instead of validation loss yields a task-specific $\varphi_{\mathrm{axis}}$, which quantifies how much one recurrence is worth on a particular task type. We evaluate this task-specific formulation on the four downstream axes described in Appendix~\ref{app:eval_suite}. For parametric knowledge and reading comprehension, $\varphi_{\mathrm{axis}}$ remains below $1$ but is higher than the validation-loss exponent, indicating a smaller parameter-sharing cost (Table~\ref{tab:downstream_optimum}). The downstream case-study fits also confirm the validation-loss findings and preserve the same relative ordering: truncated BPTT lowers $\varphi_{\mathrm{axis}}$, while hyperconnections raise it (Table~\ref{tab:downstream-case-studies}). Math word problems and reasoning primitives show no consistent differences beyond run-to-run variation at our compute budgets, including the larger extrapolation runs. The relationship between $\varphi$ and reasoning performance therefore remains unresolved. A reasoning-heavy pretraining mix might make reasoning differences between the architectures measurable at our budgets~\citep{knupp2026dreamer}.

\paragraph{Limitations.}
Our iso-depth setup limits recurrences at $r_\text{max}{=}16$. The $r^\varphi$ form is a pre-saturation local approximation valid in this range and does not include the architectural ceiling. The joint law also assumes that each additional recurrence must either consistently help ($\varphi > 1$) or consistently hurt ($\varphi < 1$) compared to the non-looped baseline. When some $r$ outperform the baseline and others fall below it, no single $\varphi$ captures both directions. The non-looped baseline is therefore an important calibration anchor for $\varphi$. $\varphi$ obtained with different baselines are not comparable. Thus, we tune all four architectures with the same recipe in Section~\ref{sec:experimental-details}, with Appendix~\ref{app:hparams} confirming LR optima agree.

\section{Conclusion}
\label{sec:conclusion}

We measure the parameter-sharing cost of looped language models as a single number, the recurrence-equivalence exponent $\varphi$. On our baseline recipe we obtain $\varphi {=} 0.46$, so each shared recurrence is worth roughly half a unique block. The fitted $\varphi$ responds to design choices, and our two case studies move it in opposite directions even though both lower validation loss. Hyperconnections raise $\varphi$ to $0.65$, while truncated BPTT lowers it to $0.38$ by re-allocating compute toward more tokens and a wider model. Raw validation loss does not separate these two cases, but $\varphi$ does. $\varphi$ additionally reflects inference cost via the compute-optimal allocation, as $\Delta\varphi > 0$ narrows the compute-optimum and lowers per-token inference FLOPs, while $\Delta\varphi < 0$ widens it. We therefore propose $\Delta\varphi$ as a metric alongside validation loss for any new looped LM recipe or architecture.

\begin{ack}
KS and GK received support from the Bavarian Collaborative Research Project PRIPREKI of the Free State of Bavaria Funding Programme “Artificial Intelligence – Data Science”.

The authors gratefully acknowledge the scientific support and resources of the AI service infrastructure LRZ AI Systems provided by the Leibniz Supercomputing Centre (LRZ) of the Bavarian Academy of Sciences and Humanities (BAdW), funded by the Bayerisches Staatsministerium für Wissenschaft und Kunst (StMWK), primarily through the BayernKI initiative.

We acknowledge EuroHPC Joint Undertaking for awarding us access to MareNostrum5 at BSC, Spain.

The authors declare no competing interests.
\end{ack}

\bibliographystyle{unsrtnat}
\bibliography{references}

\appendix

\newpage
\section{Extended Related Work}
\label{app:extended_related}

\paragraph{Scaling laws.}
\citet{kaplan2020scaling} established power-law relations between loss, model size, and training tokens. \citet{hoffmann2022training} refined the allocation (Chinchilla) and found that compute-optimal training scales parameters and tokens at roughly equal rates with compute. Subsequent work has examined learning-rate transfer~\citep{yang2021mup,ren2026muonh} and inference-aware scaling that trades training tokens for inference cost~\citep{sardana2024chinchilla, roberts2026testtimescaling}. We extend these analyses to looped architectures.

\paragraph{Iso-parameter scaling law.}
Concurrent work by~\citet{prairie2026parcae} fits an iso-parameter scaling law at fixed unique parameter count $N$. Iso-parameter and iso-depth formulations impose different inference constraints. For example, at width $s{=}34$ and 4k context length, iso-depth scaling reduces weight memory from $2.55$GB to $0.98$GB at $r{=}8$, while per-token FLOPs vary by less than $5\%$ and KV-cache memory remains constant. The corresponding iso-parameter model retains approximately same weight memory across all architectures but executes $6.6\times$ as many layers, uses $6.4\times$ the per-token FLOPs, and requires $6.6\times$ the KV-cache memory. Appendix~\ref{app:inference-cost} reports the full comparison.

Additionally, three methodological details affect direct comparison. (1)~\citet{prairie2026parcae} sample per-step recurrence counts from a distribution with mean $\mu_\text{rec}$, while we fix $r$ architecturally. (2)~They use truncated BPTT, which reduces training FLOPs. We train our main grid under full BPTT to keep training and inference FLOPs aligned with the matched non-looped baseline, and probe the truncated-BPTT alternative separately (Section~\ref{sec:case_truncbptt} shows it lowers $\varphi$). (3)~Their default input injection is diagonal ($\mathcal{O}(d)$ parameters). Ours is a linear map $W_\text{inject} \in \mathbb{R}^{d \times 2d}$ with a small FLOPs cost. Their diagonal-injection layer remains untested in our framework. Overall, the two scaling laws are complementary and answer different questions.

A scaling law that matches parameters, FLOPs, and memory simultaneously remains an open direction. \citet{knupp2026dreamer} achieve such matching, introducing FLOP-neutral capacity through sparse MoEs alongside depth attention and per-recurrence expert routing layered onto every block. Extending their method to a scaling law setup would be interesting future work.

\paragraph{Depth-adaptive inference.} A main promise of looped LMs is depth-adaptive inference. Tokens in a sequence can exit after different numbers of recurrences, spending less compute on easy predictions and more on difficult ones~\citep{dehghani2019universal, fu2025thinkathard}. Realising this requires additional choices during both training and inference. Learned exit gates add training FLOPs~\citep{zhu2025scaling}, while recurrence-count sampling makes training compute variable and complicates FLOP accounting~\citep{geiping2025scaling}. At inference, variable exit depths break uniform batching and require specialised scheduling and KV-cache handling~\citep{bae2025relaxed,bae2025mor}. Continuous depth batching provides these mechanisms and enables wall-clock speedups~\citep{schwethelm2026cdb}. Predetermined routing schemes such as Mixture-of-Recursions~\citep{bae2025mor} and LoopFormer~\citep{jeddi2026loopformer} provide an alternative with more uniform execution. A scaling law that jointly accounts for recurrence value, gate training, variable inference compute, and serving efficiency is an important direction for future work.

\paragraph{Test-time loop extrapolation.}
The optimistic picture from synthetic algorithmic tasks~\citep{fan2024looped}, of training at low recurrence counts and extrapolating to greater depth at inference, has not transferred cleanly to general language modelling. \citet{geiping2025scaling, prairie2026parcae} train their models with sampled recurrence counts extending to large values, to enable test-time scaling. However, \citet{prairie2026parcae} fit a joint training–inference scaling law whose test-time component is a saturating exponential $\mathcal{L}(T) = \mathcal{L}_\infty + Z \exp(-z T / \mu_\text{rec})$ that plateaus at $T \approx \mu_\text{rec}$. Thus, the mean recurrence in training caps the test-time frontier. Ouro~\citep{zhu2025scaling} was trained with an anytime-prediction gate and reports no inference-time gains beyond the trained depth either. Taken together, effective inference depth in trained looped LMs concentrates near the training depth distribution rather than extrapolating freely past it. We therefore treat $r$ as an architectural, not a test-time, scaling axis. Until depth extrapolation works, it does not raise the worth of a recurrence beyond what $\varphi$ captures at training time.

\section{Compute Resources}
\label{app:compute}

Experiments ran on a mix of A100-80GB and H100-80GB GPUs. The 116-run main grid, two case studies, and the effective-depth study account for the bulk of the budget. The $s{=}34$ extrapolation pair at $47$B tokens also added substantial cost. The full project, including exploratory configurations, failed runs, and side experiments on the same accounts, consumed approximately $5{,}000$ GPU-hours.

\section{Model Architecture}
\label{app:architecture}

\subsection{Implementation Details}
\label{app:architecture-details}

All architectures are decoder-only transformers built on top of nanochat~\citep{karpathy2025nanochat}, using the same pre-norm block summarised in Table~\ref{tab:arch-spec}. The layer partition for each $r$ is $\ell_\text{prelude} + r \cdot \ell_\text{recur} + \ell_\text{coda} = 20$ with $(\ell_\text{prelude}, \ell_\text{coda}) = (2, 2)$ for $r > 1$, so $\ell_\text{recur} = 16/r$ evaluates to $\{8, 4, 2\}$ for $r \in \{2, 4, 8\}$. Figure~\ref{fig:architecture-overview} right visualises the four architectures.

\begin{table}[tb]
\centering
\small
\caption{Transformer architecture.}
\label{tab:arch-spec}
\begin{tabular}{@{}ll@{}}
\toprule
Component & Details \\
\midrule
Sequence length   & $T = 2048$ tokens \\
Attention         & Full causal self-attention, no sliding window, no dropout \\
Attention backend & FlashAttention-2~\citep{dao2024flashattention2} on A100, FlashAttention-3~\citep{shah2024flashattention3} on H100 \\
Position encoding & Rotary embeddings~\citep{su2023roformer}, base $\theta=10{,}000$ \\
Head dimension    & $d_\text{head} = 128$, $n_\text{head} = d_\text{model}/128$ \\
QK normalisation  & Functional RMSNorm on $q$, $k$ before attention~\citep{dehghani2023scaling} \\
MLP activation    & Squared ReLU~\citep{so2021primer}, hidden dim $= 4 d_\text{model}$ \\
Normalisation     & Learnable RMSNorm~\citep{zhang2019rms}, pre-norm, $\epsilon{=}10^{-6}$ \\
Biases            & None (all linear layers bias-free) \\
Embeddings        & Untied \texttt{wte} and \texttt{lm\_head}, token embeddings cast to bf16 \\
Vocabulary        & Llama~2 tokenizer~\citep{touvron2023llama2}, padded $32008 \to 32064$  \\
Logit softcap     & $z = 15 \cdot \tanh(\text{logits}/15)$~\citep{gemmateam2024gemma2}, applied in fp32 before the loss \\
Dropout           & None \\
\bottomrule
\end{tabular}
\end{table}

Each transformer block computes
\begin{align*}
    \hat{x} &= x + \text{Attn}(\text{RMSNorm}(x)),
    &
    x_\text{out} &= \hat{x} + \text{MLP}(\text{RMSNorm}(\hat{x})).
\end{align*}

In addition to the pre-norms inside each block, three model-level RMSNorms are applied on the residual stream: one after the token embedding, one at the end of every recurrence iteration (so the state handed to the next iteration or to the coda has controlled scale), and one before the \texttt{lm\_head}.

\paragraph{Input injection (looped only).} For each looped architecture ($r > 1$), every recurrence iteration begins with a linear injection step
\begin{equation}
  u^{(t)} = W_\text{inject}\,[e \,\|\, h^{(t)}],
  \qquad
  W_\text{inject} \in \mathbb{R}^{d \times 2d},
  \label{eq:injection}
\end{equation}
where $e$ is the prelude output (constant across recurrences), $h^{(t)}$ is the recurrent state at iteration $t$ with $h^{(0)} = e$, and $W_\text{inject}$ is initialised as $[I \,\|\, 0]$ so that $u^{(0)} \approx e$ at the start of training. Appendix~\ref{app:injection_ablation} ablates this choice against additive-residual and no-injection variants.

\paragraph{Initialisation.}
Token embeddings are drawn from $\mathcal{N}(0, 1)$ and then cast to bf16. The LM head is $\mathcal{N}(0, 10^{-3})$. Attention and MLP weights use $\mathcal{U}(-a, a)$ with $a = \sqrt{3}/\sqrt{d_\text{model}}$ (equivalently, the same standard deviation $1/\sqrt{d_\text{model}}$ as the matched normal but with bounded tails), except \texttt{mlp.c\_proj} which uses $a = \sqrt{3}/\sqrt{4 d_\text{model}}$ (Kaiming fan-in~\citep{he2015delving} over its input width $4 d_\text{model}$). The injection layer is initialised as $[I \,\|\, 0]$, and all RMSNorm scales are initialised to one.

\paragraph{Relation to Huginn.}
We adopt the prelude-recur-coda architecture and learned concatenation-based input injection of Huginn~\citep{geiping2025scaling}. Huginn additionally samples recurrence counts during training, initialises the recurrent state randomly from $\mathcal{N}(0, 1)$, and uses truncated BPTT. The first two choices primarily support adaptive inference: recurrence-count sampling trains the model to predict after different numbers of loops, while random state initialisation stabilises the loop when increasing the recurrence count at inference. Reliable improvements from increasing recurrence at inference remain an open question, which is outside the scope of this work (Appendix~\ref{app:extended_related}). Our goal is to isolate the parameter-sharing cost at a fixed recurrence count. We therefore avoid the added stochasticity of random state initialisation and the variable per-token training FLOPs introduced by recurrence-count sampling. We use full BPTT so that training FLOPs account for every recurrence, and study truncated BPTT separately in Section~\ref{sec:case_truncbptt}. We ablate random state initialisation below.

\subsection{Input-Injection and State-Initialisation Ablations}
\label{app:injection_ablation}

Our default injection is the linear map of Equation~\ref{eq:injection}, following the concatenation-based input injection~\citep{geiping2025scaling}. The linear map adds $2 d_\text{model}^2$ parameters and, applied once per recurrence, a non-negligible FLOPs overhead that is paid back only if it improves quality. We verify this at the reference configuration ($s{=}10$, $r{=}4$, target compute $10^{18}$~FLOPs) against two parameter-free alternatives:
\begin{itemize}
    \item \textbf{Passthrough} ($u^{(t)} = h^{(t)}$): no injection, recurrence is depth-only with $h^{(0)}$ initialised from the prelude output.
    \item \textbf{Additive} ($u^{(t)} = h^{(t)} + e$): parameter-free residual injection with $h^{(0)} = 0$, so the first iteration sees $u^{(0)} = e$.
\end{itemize}
We additionally replace the default initialisation $h^{(0)}{=}e$ with a random initialisation $h^{(0)} {\sim }\mathcal{N}(0, 1)$~\citep{geiping2025scaling}, retaining the learned linear injection. All variants use the same target FLOPs budget, so the parameter-free alternatives train on more tokens than the learned-injection variants (973M vs.\ 955M, a ${\sim}2\%$ data advantage from the saved injection FLOPs). Results are summarised in Table~\ref{tab:injection_ablation}. Passthrough fails to train, showing that some form of injection is essential at this scale. Additive is competitive but trails the linear injection by $0.004$~nats despite its token advantage. Random state initialisation has little effect, trailing the deterministic default by $0.011$~nats. We therefore adopt the linear injection with deterministic state initialisation for all reported scaling-law runs. Its FLOPs overhead is accounted for in $n_i$ in Equation~\ref{eq:forward-flops} and thus included in every iso-FLOPs comparison.

Hyperconnections~\citep{zhu2025hyperconnections} are only listed for reference. We did not adopt them in the main scaling-law grid. They are used in Section~\ref{sec:case_hyperconnections} to test whether they improve $\varphi$.

\begin{table}[tb]
\centering
\caption{Input-injection ablations at $s{=}10$, $r{=}4$, and $C {=} 10^{18}$~FLOPs. Tokens differ across variants because the parameter-free and hyperconnection methods save the linear injection's FLOPs overhead.}
\label{tab:injection_ablation}
\small
\begin{tabular}{@{}lrr@{}}
\toprule
Variant & Tokens trained & Val.\ loss (nats) \\
\midrule
Linear (default)              & $955$M & $2.793$ \\
Linear + random state         & $955$M & $2.804$ \\
Additive                      & $973$M & $2.797$ \\
Passthrough                   & $973$M & $7.400$ \\
\midrule
Hyperconnections ($K{=}2$)    & $973$M & $2.757$ \\
\bottomrule
\end{tabular}
\end{table}

\section{Inference Cost under Iso-Depth and Iso-Parameter Scaling}
\label{app:inference-cost}

We compare the inference cost of iso-depth and iso-parameter loop scaling at width $s{=}34$. Iso-depth reduces the number of layers in the recurrent block as $r$ increases, keeping the effective depth fixed (Figure~\ref{fig:architecture-overview}). Iso-parameter scaling retains the full recurrent block, keeping the unique parameter count approximately fixed while effective depth grows with $r$. Table~\ref{tab:inference-cost} reports the resulting inference cost of both variants based on weight memory, per-token FLOPs, KV-cache memory, and the maximum number of concurrent sequences that fit on an 80GB accelerator. Under iso-depth scaling, weight memory decreases from $2.55$GB to $0.98$GB, while per-token FLOPs vary by less than $5\%$ and the KV-cache footprint remains constant. Under iso-parameter scaling, the $r{=}8$ model executes $6.6\times$ as many layers as the baseline, uses $6.4\times$ the FLOPs per generated token, and supports 16 instead of 105 concurrent sequences at a 4k-token context length. Iso-depth therefore isolates the parameter-sharing trade-off while controlling the inference costs that grow with recurrence. Iso-depth comparisons can also be constructed at greater effective depths. For example, the 132 effective layers of the iso-parameter $r{=}8$ model could be matched by a 132-layer non-looped baseline.

\begin{table}[tb]
\centering
\caption{Inference costs at width $s{=}34$ ($d_\text{model}{=}2176$) in bf16.
$N$ excludes embedding parameters, weight memory includes them.
FLOPs are calculated per generated token at the stated context length, including attention and the output projection.
The sequence counts assume an 80GB accelerator with 2GB reserved for activations and other workspace.
The looped iso-parameter models match the baseline parameter count up to the additional 10M parameters of the input-injection layer.}
\label{tab:inference-cost}
\scriptsize
\setlength{\tabcolsep}{2.5pt}
\resizebox{\textwidth}{!}{%
\begin{tabular}{@{}llrrr|rrr|rrr@{}}
\toprule
 & & & & & \multicolumn{3}{c|}{4k context} & \multicolumn{3}{c}{16k context} \\
Model & Design & $\ell_\text{eff}$ & $N$ (M) & Weights (GB) & GFLOPs/tok. & KV/seq.\ (GB) & Seqs. & GFLOPs/tok. & KV/seq.\ (GB) & Seqs. \\
\midrule
$r{=}1$ & Baseline  & 20  & 1136 & 2.55 &  3.13 & 0.71 & 105 &  5.26 &  2.85 & 26 \\
\midrule
$r{=}2$ & Iso-depth & 20  &  691 & 1.66 &  3.16 & 0.71 & 107 &  5.30 &  2.85 & 26 \\
$r{=}4$ & Iso-depth & 20  &  464 & 1.21 &  3.20 & 0.71 & 107 &  5.34 &  2.85 & 26 \\
$r{=}8$ & Iso-depth & 20  &  350 & 0.98 &  3.28 & 0.71 & 108 &  5.42 &  2.85 & 27 \\
\midrule
$r{=}2$ & Iso-param. & 36  & 1146 & 2.57 &  5.55 & 1.28 &  58 &  9.40 &  5.13 & 14 \\
$r{=}4$ & Iso-param. & 68  & 1146 & 2.57 & 10.37 & 2.42 &  31 & 17.64 &  9.70 &  7 \\
$r{=}8$ & Iso-param. & 132 & 1146 & 2.57 & 20.00 & 4.71 &  16 & 34.12 & 18.82 &  4 \\
\bottomrule
\end{tabular}%
}
\end{table}

\section{Hyperparameter Tuning}
\label{app:hparams}

\subsection{Learning Rate Sweep}
\label{app:lr-sweep}

We sweep the MuonH~\citep{ren2026muonh, wen2025hyperball} matrix learning rate at $s{=}10$ with a tokens-per-parameter ratio of 10 (${\sim}1$B training tokens) and batch size $B = 262{,}144$ (256K), independently for each architecture, using eight LR values per architecture in the range $\eta \in [0.008, 0.024]$ with extra density around the optimum. The batch size was chosen from a separate sweep at the same reference configuration over $B \in \{256\text{K}, 512\text{K}, 1\text{M}\}$ tokens, where $256$K yielded uniformly lower loss for both architectures.

\begin{figure}[tb]
    \centering
    \includegraphics[width=0.65\textwidth]{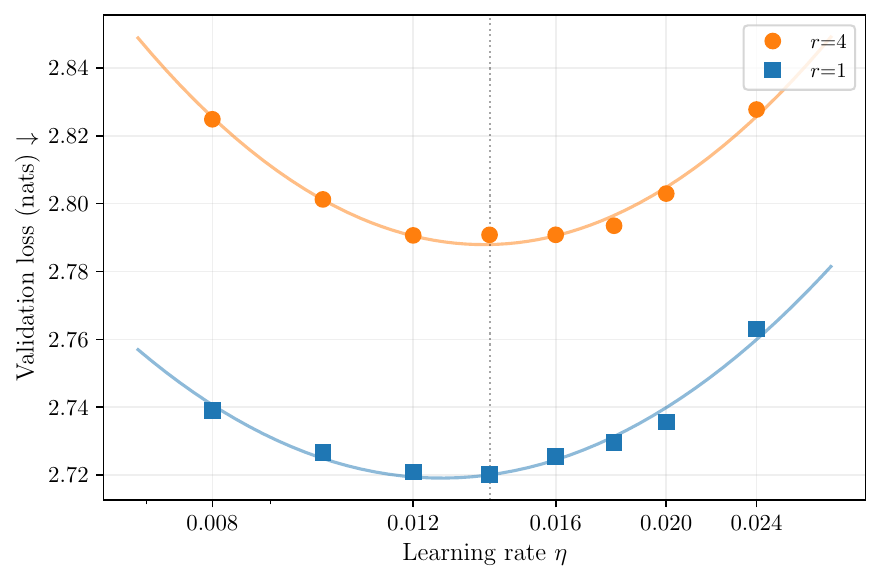}
    \caption{Learning rate sweep at $s{=}10$ (ratio 10, $B = 256$K). Both architectures exhibit a clear U-shaped loss landscape with a shared optimum near $\eta^* \approx 0.014$. The dotted vertical line marks $\eta = 0.014$, the base LR adopted for all scaling-law runs.}
    \label{fig:lr-sweep}
\end{figure}

Both architectures converge to a shared optimum near $\eta^* \approx 0.014$, which we adopt as the base learning rate for all subsequent experiments (Figure~\ref{fig:lr-sweep}).

\subsection{Transfer Validation}
\label{app:transfer}

Under the HyperP framework~\citep{ren2026muonh} the \emph{base} learning rate $\eta_\text{base}$ (the value fed into HyperP before its width and data corrections are applied) should be invariant to both width and training horizon (after the $D^{-0.32}$ data-scaling correction of the HyperP LR rule, with $D$ the training-token count). We verify both claims by repeating the LR sweep under varied conditions and measuring the regret: the loss penalty of using $\eta_\text{base} = 0.014$ instead of the per-condition optimum.

\begin{figure}[t]
    \centering
    \includegraphics[width=\textwidth]{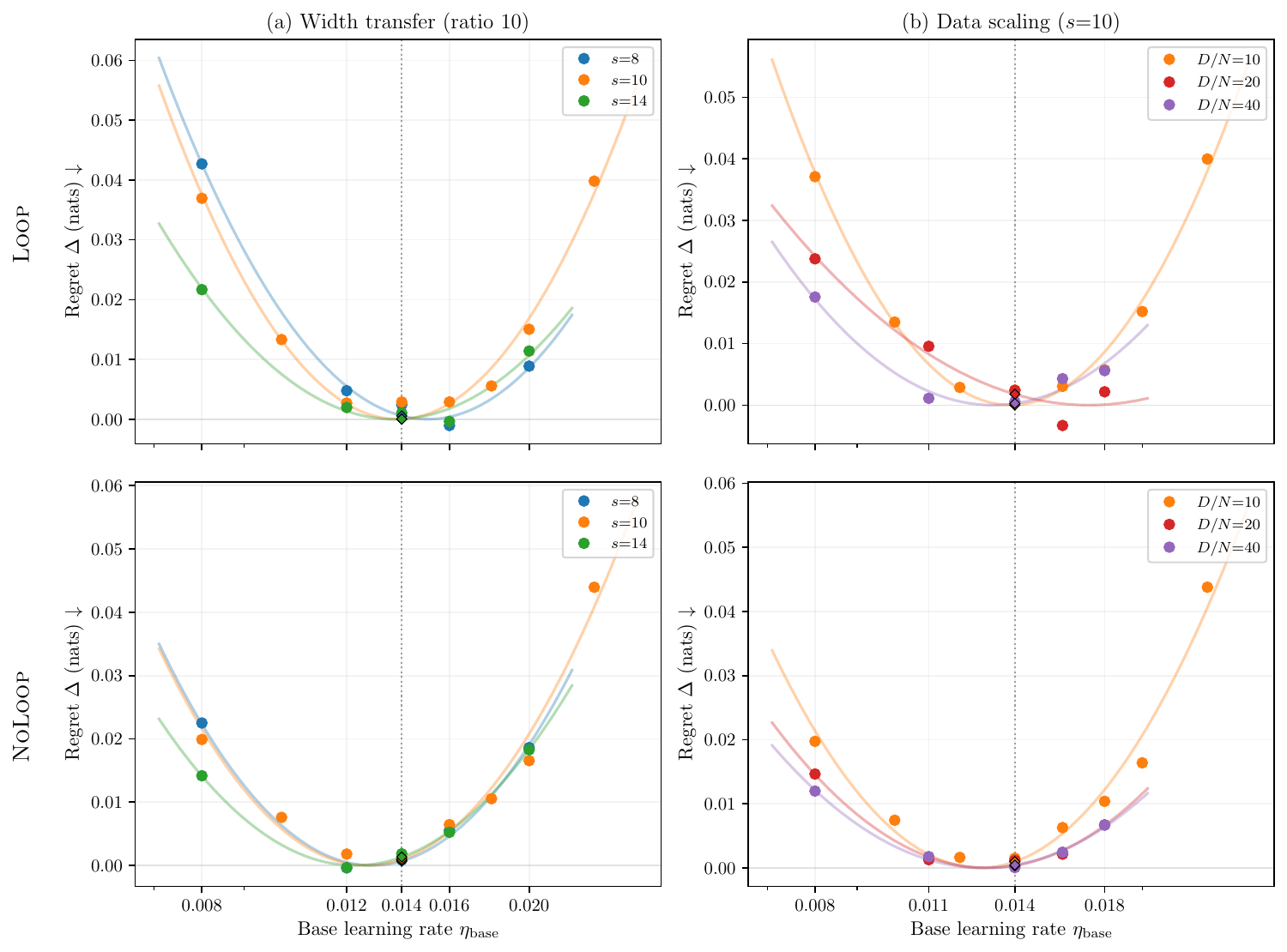}
    \caption{Transfer validation. Regret (loss above the per-condition optimum) vs.\ base learning rate $\eta_\text{base}$. Vertical dotted line marks $\eta_\text{base} = 0.014$; diamond markers show the regret at that point. Rows split by architecture (looped $r{=}4$, top; non-looped $r{=}1$, bottom). All conditions incur $\leq 0.005$~nats regret, so $\eta_\text{base} = 0.014$ transfers cleanly across width and training horizon.}
    \label{fig:transfer}
\end{figure}

\paragraph{Width transfer.}
We sweep at $s \in \{8, 10, 14\}$ (ratio 10, $B = 256$K). Figure~\ref{fig:transfer} (left column) shows the regret U-curves in base LR space: all minima cluster near $\eta_\text{base} = 0.014$ with a maximum regret of $0.004$~nats ($s{=}8$ looped). As a lightweight sanity check past the sweep range, we run the two candidate LRs $\eta_\text{base} \in \{0.012, 0.014\}$ at $s{=}18$ for the looped architecture and find $0.014$ marginally better ($2.473$ vs.\ $2.476$~nats), confirming that $0.014$ remains near-optimal.

\paragraph{Data scaling.}
We sweep at $s{=}10$ ($B = 256$K) with ratios $\{10, 20, 40\}$, spanning a $4\times$ range in training tokens. If the $T^{-0.32}$ exponent is correct, the data-scaling correction adjusts the effective LR automatically and the optimal base LR should remain constant. Figure~\ref{fig:transfer} (right column) confirms this: regret at $\eta_\text{base} = 0.014$ stays below $0.005$~nats across all ratios for both architectures.

\section{Model Sizes and Trained Tokens}
\label{app:grid}

For each compute budget we sweep model width to find the compute-optimal point at every recurrence count $r \in \{1, 2, 4, 8\}$. Table~\ref{tab:grid} reports the unique non-embedding parameter count $N(s, r)$ at each width and the training tokens $D$ used at each (budget, width) cell. Looped models train on slightly fewer tokens due to input injection compute overhead. Empty cells are untested widths.

\begin{table}[tb]
\centering
\caption{Model sizes and trained tokens. Left: unique non-embedding parameter count $N$ (M) per (width, recurrence) cell. Right: training tokens (B) per (width, budget) cell for $r=1$.}
\label{tab:grid}
\small
\begin{tabular}{@{}c|rrrr|rrrrrr@{}}
\toprule
 & \multicolumn{4}{c|}{Unique params $N$ (M)} & \multicolumn{6}{c}{Training tokens (B) per budget (FLOPs)} \\
$s$ & $r{=}1$ & $r{=}2$ & $r{=}4$ & $r{=}8$ &
   $4.64{\cdot}10^{17}$ & $10^{18}$ & $2.15{\cdot}10^{18}$ &
   $4.64{\cdot}10^{18}$ & $10^{19}$ & $2.15{\cdot}10^{19}$ \\
\midrule
 6 &  35.4 & 21.5 & 14.5 & 10.9 & 0.98 & 2.10 & ---  & ---  & ---  & --- \\
 8 &  62.9 & 38.3 & 25.7 & 19.4 & 0.64 & 1.36 & 2.95 & ---  & ---  & --- \\
10 &  98.3 & 59.8 & 40.2 & 30.3 & 0.45 & 0.97 & 2.08 & 4.49 & ---  & --- \\
12 & 141.6 & 86.1 & 57.8 & 43.7 & 0.34 & 0.72 & 1.55 & 3.34 & 7.13 & --- \\
14 & 192.7 &117.2 & 78.7 & 59.4 & ---  & 0.56 & 1.20 & 2.59 & ---  & --- \\
16 & 251.7 &153.1 &102.8 & 77.6 & ---  & ---  & 0.96 & 2.07 & 4.43 & --- \\
18 & 318.6 &193.8 &130.1 & 98.2 & ---  & ---  & ---  & 1.70 & 3.62 & 7.78 \\
20 & 393.3 &239.2 &160.6 &121.3 & ---  & ---  & ---  & 1.42 & 3.05 & --- \\
24 & 566.3 &344.5 &231.2 &174.6 & ---  & ---  & ---  & ---  & 2.20 & 4.71 \\
28 & 770.8 & 468.9 &314.7 & 237.7 & ---  & ---  & ---  & ---  & ---  & 3.58 \\
34 &1136.5 & 691.4 &464.1 & 350.4 & ---  & ---  & ---  & ---  & ---  & 2.52 \\
\bottomrule
\end{tabular}
\end{table}

\begin{figure}[tb]
    \centering
    \includegraphics[width=\textwidth]{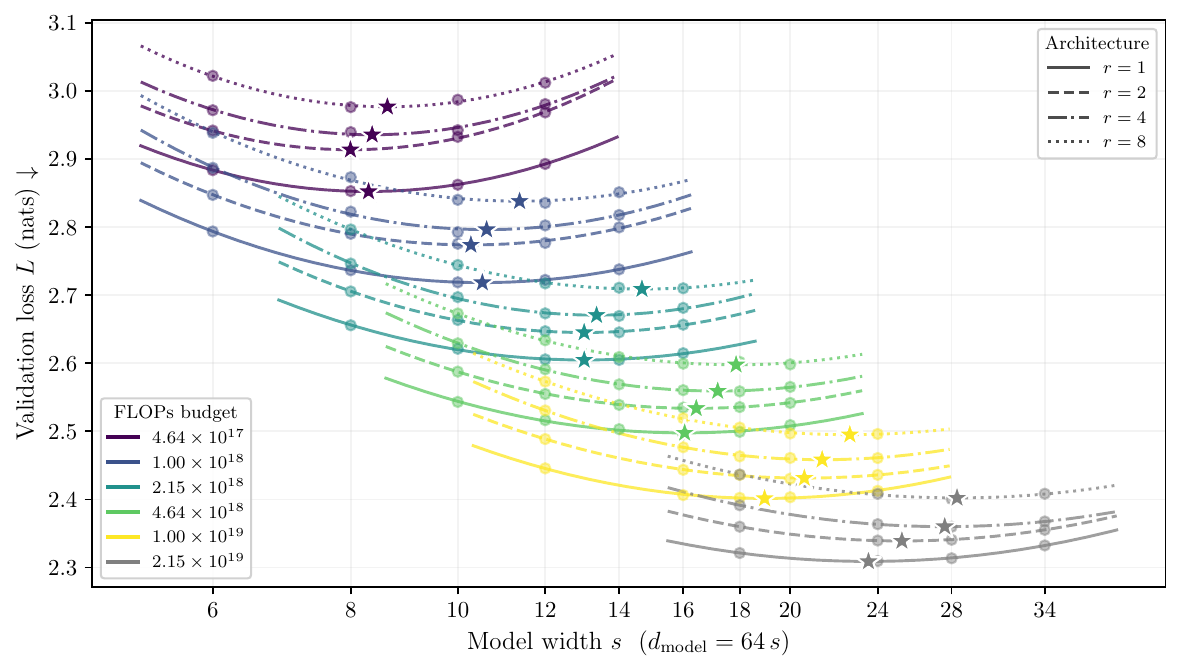}
    \caption{Scaling curves at fixed compute budgets, against model width $s$. Same runs as Figure~\ref{fig:isoflops}. Thin curves are per-(budget, $r$) parabolic fits in $\log s$. Stars mark the fitted compute-optimal $(s^*, L^*)$ points. At matched width, looped models reach higher loss than the non-looped baseline and their optima sit mostly at larger $s$.}
    \label{fig:isoflops-width}
\end{figure}

\section{Scaling Law Fit Diagnostics}
\label{app:fit-residuals}

We conduct robustness checks for the per-architecture Chinchilla fits (Equation~\ref{eq:chinchilla}) and the joint $(N_\text{once}, N_\text{rec}, D, r)$ law (Equation~\ref{eq:joint-3d}). Residuals for the per-architecture fits, aggregate residual statistics for the joint fit, the block-bootstrap procedure behind the $\varphi$ confidence interval, and stability of $\varphi$ across budget halves.

\paragraph{Optimisation details.}
The Huber-on-log objective of Equation~\ref{eq:huber-obj} is non-convex, so for both the per-architecture and joint fits we take the best of $500$ random L-BFGS-B restarts. Parameters are constrained to the box $a, b \in [-5, 35]$, $\alpha, \beta \in [0, 2.5]$, $e \in [-3, 2]$ (and $\varphi \in [-3, 3]$ for the joint fit), with starting points drawn uniformly inside the box and a per-restart cap of $10{,}000$ iterations.

\subsection{Per-Architecture and Joint Fit Residuals}

\begin{figure}[tb]
    \centering
    \includegraphics[width=\textwidth]{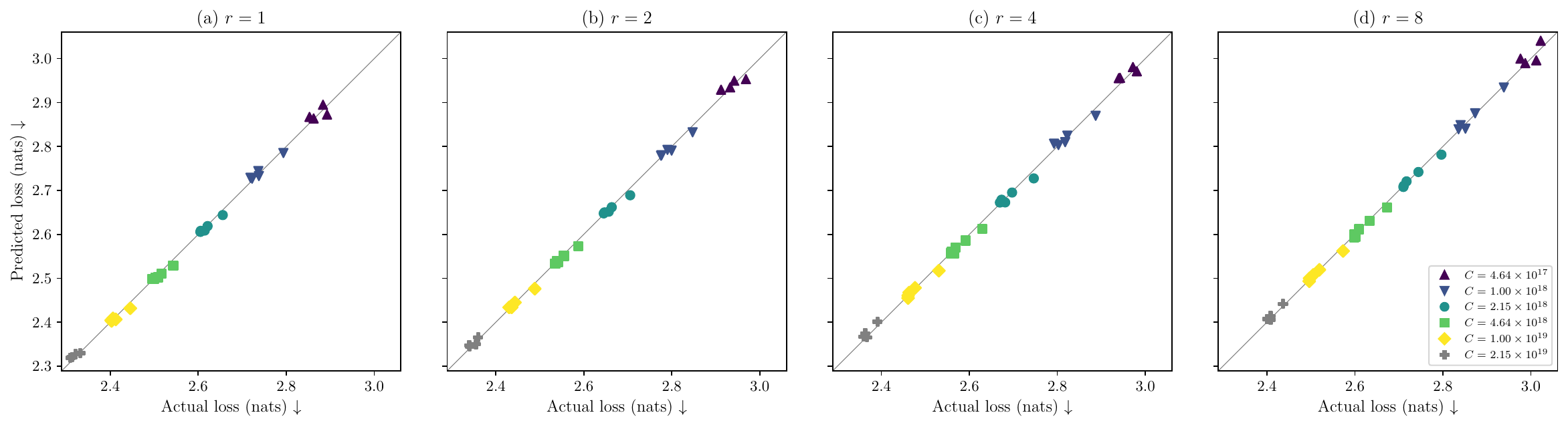}
    \caption{Per-architecture Chinchilla fit quality: predicted vs.\ actual validation loss, one panel per architecture ($r \in \{1, 2, 4, 8\}$). Markers redundantly encode the compute budget by both shape and colour. Points cluster tightly around the identity line across all four architectures.}
    \label{fig:fit-quality}
\end{figure}

\begin{figure}[tb]
    \centering
    \includegraphics[width=\textwidth]{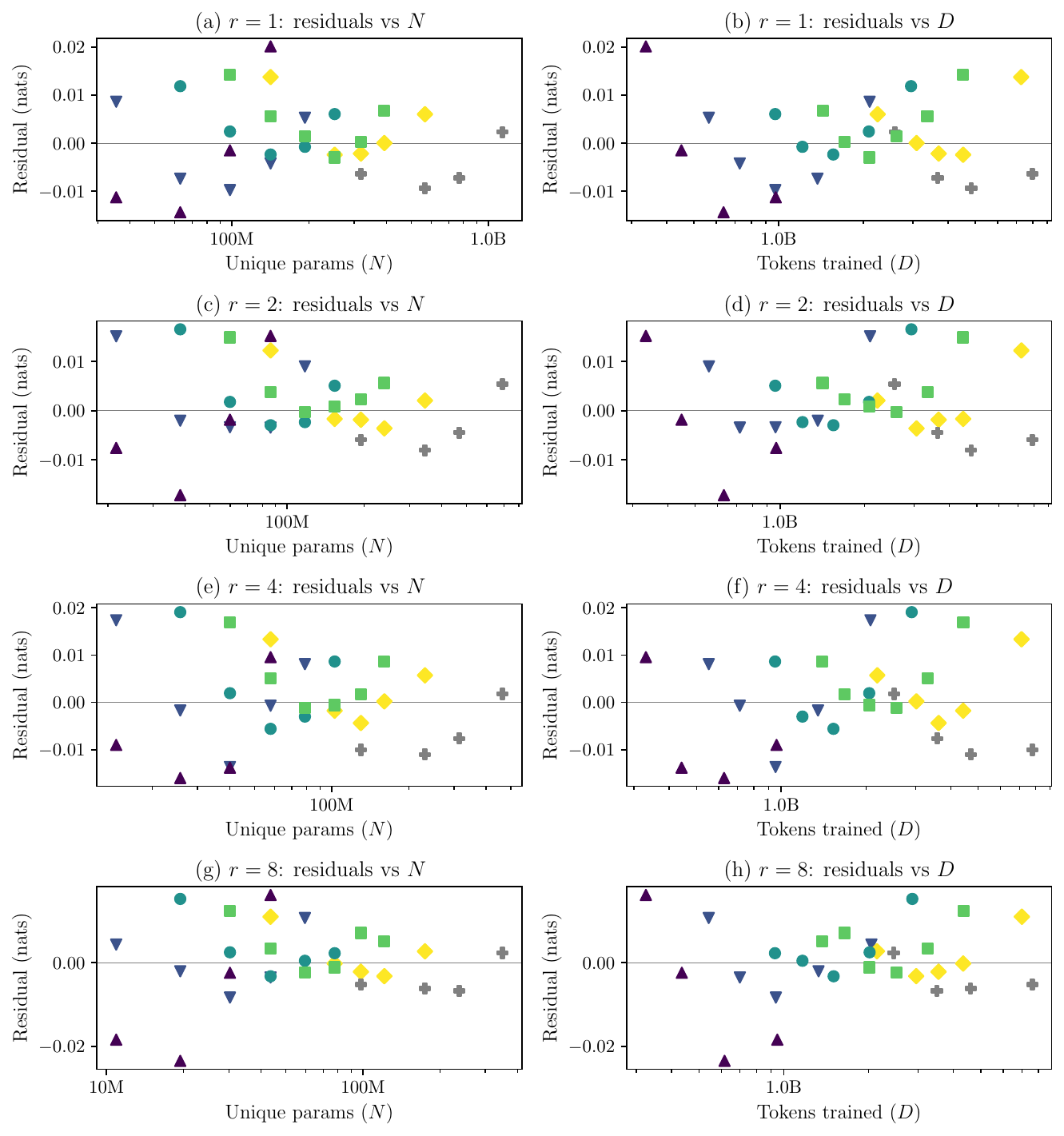}
    \caption{Per-architecture Chinchilla fit residuals (actual $-$ predicted) vs.\ unique parameters $N$ (left column) and tokens $D$ (right column). Rows are the four architectures $r \in \{1, 2, 4, 8\}$. Markers encode the compute budget by both shape and colour.}
    \label{fig:fit-residuals}
\end{figure}

Figure~\ref{fig:fit-quality} shows predicted vs.\ actual validation loss for the four per-architecture Chinchilla fits. Points cluster tightly around the diagonal with a maximum residual below $0.007$~nats. Figure~\ref{fig:fit-residuals} plots the same residuals against $N$ and $D$ and shows no systematic bias across either axis.

On the joint law, which fits all 116 runs with six shared parameters $(A, \alpha, B, \beta, E, \varphi)$ rather than 20 per-architecture parameters, residuals are naturally larger but still small: max $|\text{resid}| = 0.036$~nats, pooled $\text{RMSE} = 0.010$~nats, and $R^2 = 0.997$. Table~\ref{tab:joint-residuals-per-r} breaks the joint-fit residuals down by architecture. RMSE is comparable across $r$ ($0.009$--$0.011$~nats) and per-$r$ means lie within $\pm 0.006$~nats of zero, so a single shared $\varphi$ fits all four architectures with comparable accuracy and no architecture absorbs a disproportionate share of the residual mass.

\begin{table}[tb]
\centering
\caption{Joint-law residual statistics broken down by architecture (actual $-$ predicted, in nats). Each architecture contributes 29 runs to the joint fit.}
\label{tab:joint-residuals-per-r}
\small
\begin{tabular}{@{}lrrrr@{}}
\toprule
Architecture & $n$ & mean resid & max $|\text{resid}|$ & RMSE \\
\midrule
$r{=}1$ & $29$ & $-0.001$ & $0.018$ & $0.009$ \\
$r{=}2$ & $29$ & $+0.004$ & $0.036$ & $0.011$ \\
$r{=}4$ & $29$ & $-0.006$ & $0.023$ & $0.011$ \\
$r{=}8$ & $29$ & $+0.001$ & $0.029$ & $0.010$ \\
\bottomrule
\end{tabular}
\end{table}

\subsection{Bootstrap Procedure}
\label{app:bootstrap}

The $95\%$ CI reported alongside the joint-law point estimate ($\varphi = 0.46$, $[0.41, 0.53]$) is a block bootstrap over \emph{(budget, architecture) cells}: each cell groups all widths trained at a given (compute budget, $r$) pair, so resampling respects the experimental block structure rather than treating individual runs as independent. We draw 200 resamples with replacement of the non-empty cells ($6$ budgets $\times$ $4$ architectures), refit the joint law on each resample, and report the $2.5$th / $97.5$th percentiles of the resulting $\varphi$ distribution. Zero resamples reach either $\varphi = 0$ or $\varphi = 1$. We do not bootstrap restricted variants of the law.

\subsection{Stability Across Budget Halves}

Refitting the joint law separately on the low-budget half ($C \le 2.15 \times 10^{18}$, $n{=}56$ runs) gives $\varphi = 0.44$, and refitting on the high-budget half ($C \ge 4.64 \times 10^{18}$, $n{=}60$ runs) gives $\varphi = 0.49$. $\varphi$ therefore does not drift with scale inside our compute window, and the bootstrap CI comfortably contains both half-window estimates.

\subsection{Varying Effective Depth}
\label{app:depth-allocation}

Our primary scaling sweep varies model width while fixing the effective depth at 20 layers. The iso-depth setting only requires architectures compared at the same model width to have equal effective depth, it does not require different model widths to use the same depth. We therefore test whether our main findings depend on the fixed 20-layer design by repeating the $r \in \{1,2,4,8\}$ width sweep at the four smallest compute budgets with effective depths of 18 and 24 layers. Changing the effective depth of a looped model also requires choosing how layers are allocated between the prelude, recurrent block, and coda. We keep the recurrent block's contribution to effective depth fixed at $r\ell_\text{recur}=16$ and set $(\ell_\text{prelude}, \ell_\text{coda})$ to $(1,1)$, $(2,2)$, and $(4,4)$ for the 18-, 20-, and 24-layer configurations, respectively. Each configuration includes a depth-matched non-looped baseline. The results change little across the three effective depths. Pooling all 240 runs gives a shared recurrence-equivalence exponent of $\varphi=0.468$ with a bootstrap 95\% CI of $[0.442,0.494]$, close to the primary estimate of $0.46$.

\subsection{Example: Equivalent Model Sizes at $r{=}4$}
\label{app:equiv-sizes}

We compare a looped $r{=}4$ model and a non-looped baseline at the same width and effective depth. The resulting unique-parameter and effective-parameter counts are purely architectural ratios.

Take any $r{=}1$ configuration with $N(r{=}1) = 1$B and width $d_\text{model}$. At the same $d_\text{model}$ and $\ell_\text{eff}{=}20$ with $(\ell_\text{prelude}, \ell_\text{coda}) = (2, 2)$, the $r{=}4$ variant uses $\ell_\text{recur}{=}4$ and the unique-parameter ratio (from the $N_\text{once}, N_\text{rec}$ split in Section~\ref{sec:specification}, dropping the small injection term $n_i = n_b/6$) is
\begin{equation*}
\frac{N(r{=}4)}{N(r{=}1)} = \frac{(\ell_\text{prelude} + \ell_\text{recur} + \ell_\text{coda})\,n_b}{\ell_\text{eff}\,n_b} = \frac{8}{20} = 0.40,
\end{equation*}
so the $r{=}4$ variant has $0.40 \times$ the unique parameters of the non-looped model: ${\approx}410$M including the injection term. The effective-parameter ratio under the joint law (Equation~\ref{eq:joint-3d}) is
\begin{equation*}
\frac{N_\text{eff}(r{=}4; \varphi)}{N(r{=}1)} = \frac{N_\text{once} + 4^\varphi N_\text{rec}}{\ell_\text{eff}\,n_b} \approx \frac{4 + 4^\varphi \cdot 4}{20} \stackrel{\varphi=0.46}{\approx} 0.58,
\end{equation*}
giving the ${\approx}580$M figure. Per-token training FLOPs depend only on the executed-layer count, which is identical at fixed $d_\text{model}$ up to the ${\sim}3\%$ injection overhead of Equation~\ref{eq:forward-flops}, so the $r{=}4$ model trains at the same per-step cost as the 1B non-looped baseline. The same calculation at any other reference size gives the same percentages.

\section{Case Study Details}
\label{app:probing}

\subsection{Methods}
\label{app:probing_methods}

\paragraph{Truncated backpropagation.}
We follow~\citet{prairie2026parcae} and set the gradient window to $r_\text{bwd} = \lceil r/2 \rceil$ for $r \in \{2, 4, 8\}$, giving $r_\text{bwd} \in \{1, 2, 4\}$. The forward pass is unchanged. After the $i$-th recurrence, the recurrent state $s^{(i)}$ is detached for all $i < r - r_\text{bwd}$, so gradients flow only through the last $r_\text{bwd}$ iterations. Detached iterations skip the backward pass and save roughly half the FLOPs of a full recurrent iteration, which translates to about $30\%$ fewer training FLOPs per token under our prelude-recur-coda partition,
\begin{equation}
F_\text{train}^\text{trunc}(r) = \bigl(2 (r - r_\text{bwd}) + 6 r_\text{bwd}\bigr)(\ell_\text{recur} n_b + n_i) + 6 (\ell_\text{prelude} + \ell_\text{coda}) n_b.
\label{eq:trunc-flops}
\end{equation}
The freed compute is reinvested as more tokens at fixed budget. The empirical median ratio across all (budget, $r$, $s$) cells is $D_\text{trunc} / D_\text{full} = 1.315$.

\paragraph{Hyperconnections.}
We use our own implementation of a looped transformer with hyperconnections~\citep{zhu2025hyperconnections}. Note that~\citet{zeitounHyperloopTransformers2026} concurrently proposed a similar architecture. We replace the linear input-injection layer with $K{=}2$ parallel lane states $\ell^{(i)} \in \mathbb{R}^{K \times d_\text{model}}$ mixed at every recurrence iteration. Lanes are initialised by broadcasting the prelude output $e$ across all $K$ slots, $\ell^{(0)} = (e, \ldots, e)$. At iteration $i \in \{0, \ldots, r-1\}$ the recurrent block sees
\begin{align}
u^{(i)} &= \alpha_i \cdot \ell^{(i)}, &
s^{(i)} &= \text{RecurBlock}(u^{(i)}), &
\ell^{(i+1)} &= M_i \ell^{(i)} + \beta_i \otimes s^{(i)},
\label{eq:hyperconnect}
\end{align}
with per-iteration mixing parameters $\alpha_i \in \mathbb{R}^{K}$, $M_i \in \mathbb{R}^{K \times K}$, $\beta_i \in \mathbb{R}^{K}$, for a total of $r(K^2 + 2K)$ scalars across all iterations ($32$ at $K{=}2$, $r{=}4$). The coda receives the sum-pooled lanes $\sum_k \ell^{(r)}_k$. We adopt the cyclic initialisation of~\citet{zhu2025hyperconnections}, $\alpha_i = \mathbf{e}_{i \bmod K}$, $M_i = I$, $\beta_i = \mathbf{1}$, so that the first training iteration reduces to a plain looped forward pass. All hyperconnect runs use full BPTT. The per-token training-FLOPs cost of the mixing operations is $6 r(K^2 + 2K) d_\text{model}$ (forward $2\times$ plus backward $4\times$), accounted for in our budget calculations and negligible: two orders of magnitude smaller than the linear injection it replaces.

\subsection{Fit Quality}
\label{app:probing_fitquality}

\begin{figure}[tb]
    \centering
    \includegraphics[width=\textwidth]{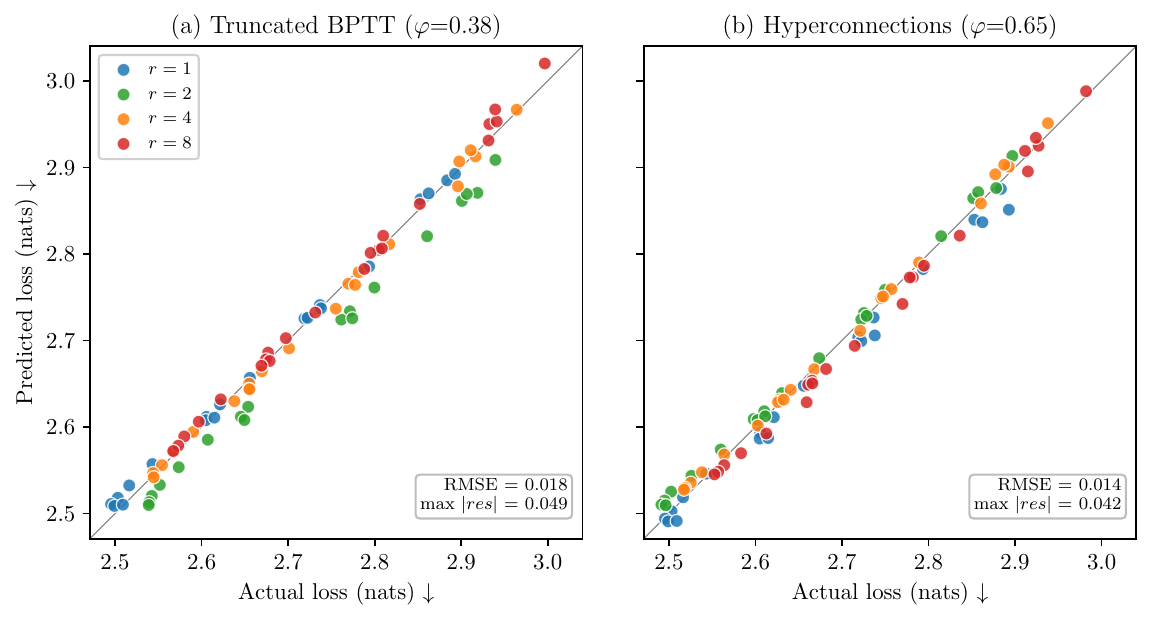}
    \caption{Case-study fit quality: predicted vs.\ actual validation loss under the joint-law refits of Table~\ref{tab:probe_fits}. Markers encode recurrence $r \in \{1, 2, 4, 8\}$ by colour. Annotation reports root-mean-square and maximum absolute residual.}
    \label{fig:probing_fit_quality}
\end{figure}

Figure~\ref{fig:probing_fit_quality} shows predicted vs.\ actual validation loss for both case-study joint fits, the analogue of Figure~\ref{fig:fit-quality} on the main grid. Points cluster on the diagonal in both panels, with overall RMSEs an order of magnitude below the inter-architecture spread.

The truncated-BPTT panel shows where the $R^2 = 0.983$ is lost. Most of the residual mass sits at $r{=}2$ ($r_\text{bwd} = 1$), where the joint law systematically under-predicts loss. Only a single recurrence receives a direct gradient, which likely undertrains the looping mechanism for that architecture. The $r \in \{4, 8\}$ rows are well-behaved. Refitting on $r \in \{4, 8\}$ alone (Table~\ref{tab:probe_fits}) raises $R^2$ from $0.983$ to $0.996$, removing the $r{=}2$ residual mass without changing $\varphi$.

The hyperconnections panel does not show a comparable architecture-specific structure. Residuals are uniformly small across $r$, and the fewer runs (83 vs.\ 116 in the main fit) and narrower compute span (four of six budgets) are the main sources of $R^2$ loss relative to the main joint fit.

\subsection{Compute-Optimal Allocation Under the Case Studies}
\label{app:probing_compute_optimal}

The shift in $\varphi$ has predictable consequences for compute-optimal allocation (Figure~\ref{fig:trunc_hc_isoflops}). Treating $r$ as fixed and writing the joint law as $L = E + (A \cdot g_r^{-\alpha}) N^{-\alpha} + B D^{-\beta}$ with $g_r = N_\text{once}/N + r^\varphi N_\text{rec}/N$, a higher $\varphi$ raises $g_r$ at $r > 1$, which lowers the effective parameter amplitude $A_\text{eff} = A g_r^{-\alpha}$, which in turn lowers the compute-optimal width $N^*(C)$ for that $r$.

\paragraph{Truncated BPTT.}
$\varphi$ falls from $0.45$ to $0.38$, so $g_r$ shrinks at every $r > 1$ and the compute-optimal width $s^*$ widens further than under full BPTT. The picture is therefore one of larger looped models, consistent with the rightward shift of the truncated BPTT compute-optimal stars in Figure~\ref{fig:trunc_hc_isoflops} (left). Wider compute-optimal models also raise per-token inference FLOPs, so the recipe trades training FLOPs for inference FLOPs at fixed deployment budget.

\paragraph{Hyperconnections.}
$\varphi$ rises from $0.45$ to $0.65$, so $g_r$ at $r > 1$ grows and the compute-optimal width $s^*$ contracts. The compute-optimal stars on the hc panel of Figure~\ref{fig:trunc_hc_isoflops} (right) sit at smaller $N^*$ than the corresponding full-BPTT linear-injection stars at matched budgets. Lower $N^*$ at the same compute also means lower per-token inference FLOPs, the converse of the truncated BPTT result.

\section{Downstream Task Evaluation}
\label{app:eval_suite}

We evaluate the trained models on downstream tasks to determine how the validation-loss gap between looped and non-looped models transfers across task types. We test whether looped LMs lose more on tasks that rely on information stored in their parameters, where their lower unique parameter counts may matter most, and whether they gain on tasks involving context use and multi-step reasoning. We study these questions across the baseline iso-FLOPs grid (Section \ref{sec:scaling}), the truncated-BPTT and hyperconnection case studies (Section \ref{sec:case_studies}), and our $s{=}34$ extrapolation run (Section~\ref{sec:scaling-fits}).

\subsection{Evaluation Protocol}
\label{app:eval_axes}

Our evaluation suite combines 18 tasks from the CORE benchmark~\citep{li2024datacomplm} and \citet{saunshi2025understanding} into four categories (Table~\ref{tab:downstream_tasks}). We retain the source benchmarks' few-shot settings, except that we reduce CoQA to 1-shot to fit our 2{,}048-token context window. Following~\citet{heineman2025signal}, we report per-token loss on the gold continuation and average the resulting task losses within each category. For compute-optimal comparisons, we select the checkpoint with the lowest validation loss for each architecture and compute budget.

\begin{table}[tb]
\centering
\caption{Downstream evaluation suite. LM and MC denote language-modelling and multiple-choice tasks. Samples gives the number of evaluated examples. TriviaQA and BigBench QA-Wikidata are uniformly subsampled to 10{,}000 examples.}
\label{tab:downstream_tasks}
\small
\begin{tabular}{@{}p{3.7cm}lrll@{}}
\toprule
Axis & Task & Samples & Shots & Type \\
\midrule
\multirow{5}{=}{Parametric knowledge\\[1pt]\footnotesize\emph{(closed-book factual questions)}}
  & TriviaQA \citep{joshi2017triviaqa}             & 10{,}000 & 5 & LM \\
  & NaturalQuestions \citep{kwiatkowski2019naturalquestions} & 3{,}610  & 5 & LM \\
  & WebQuestions \citep{berant2013webquestions}         & 2{,}032  & 5 & LM \\
  & BigBench QA-Wikidata \citep{srivastava2022bigbench} & 10{,}000 & 10 & LM \\
  & ARC-Easy \citep{clark2018arc}             & 2{,}376  & 10 & MC \\
\midrule
\multirow{5}{=}{Reading comprehension\\[1pt]\footnotesize\emph{(answers determined by a passage in the prompt)}}
  & Lambada-OpenAI \citep{paperno2016lambada}              & 5{,}153  & 0 & LM \\
  & TyDiQA-GoldP \citep{clark2020tydiqa}         & 440     & 3 & LM \\
  & SQuADv2 \citep{rajpurkar2018squadv2}              & 5{,}928  & 3 & LM \\
  & DROP \citep{dua2019drop}                 & 9{,}535  & 3 & LM \\
  & CoQA \citep{reddy2019coqa}                 & 7{,}983  & 1 & LM \\
\midrule
\multirow{3}{=}{Math word problems\\[1pt]\footnotesize\emph{(elementary-school arithmetic in words)}}
  & SVAMP \citep{patel2021svamp}                & 300     & 5 & LM \\
  & ASDiv \citep{miao2020asdiv}                & 2{,}305  & 5 & LM \\
  & MAWPS \citep{koncel2016mawps}                & 1{,}772  & 5 & LM \\
\midrule
\multirow{5}{=}{Reasoning primitives\\[1pt]\footnotesize\emph{(variable look-ups and string puzzles)}}
  & VarAssign d0 (math) \citep{saunshi2024stacking}  & 1{,}000  & 5 & LM \\
  & VarAssign d0 (code) \citep{saunshi2024stacking}  & 1{,}000  & 5 & LM \\
  & VarAssign d1 (math) \citep{saunshi2024stacking}  & 1{,}000  & 5 & LM \\
  & VarAssign d1 (code) \citep{saunshi2024stacking}  & 1{,}000  & 5 & LM \\
  & BigBench CS-algorithms \citep{srivastava2022bigbench} & 1{,}320  & 10 & LM \\
\bottomrule
\end{tabular}
\end{table}

\subsection{Baseline Grid Results}

\begin{figure}[tb]
    \centering
    \includegraphics[width=\textwidth]{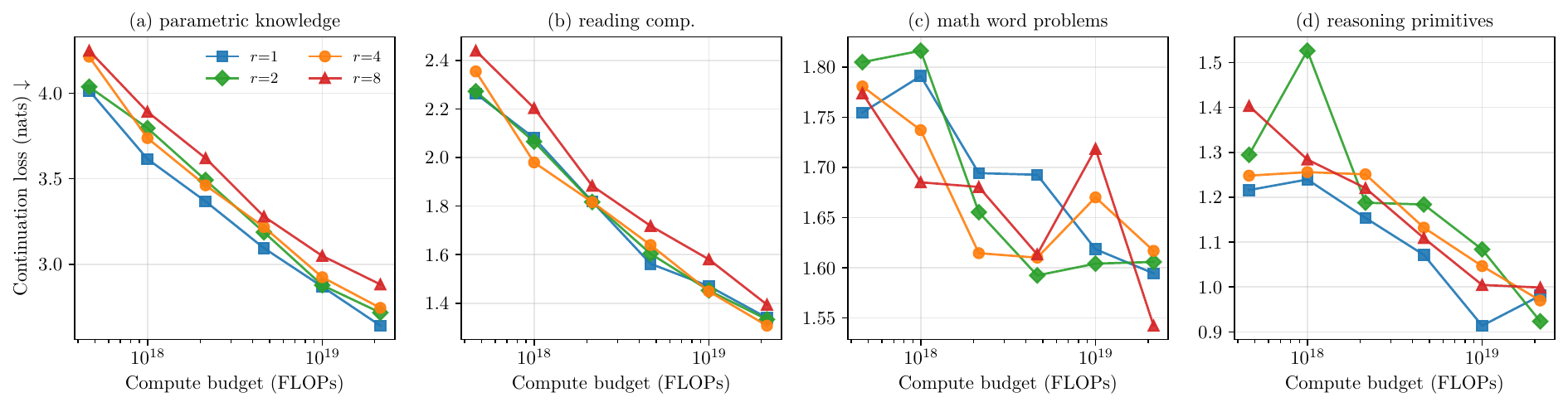}
    \caption{Downstream continuation loss across the baseline iso-FLOPs grid. Each point shows the compute-optimal model of one architecture at the corresponding compute budget. Lower is better.}
    \label{fig:downstream-scaling}
\end{figure}

 We first examine how downstream performance changes with recurrence across the baseline iso-FLOPs grid of Section \ref{sec:scaling}. Figure~\ref{fig:downstream-scaling} reports average continuation loss for the compute-optimal model of each architecture at every compute budget.

Parametric knowledge follows the validation-loss ordering most clearly. The non-looped baseline performs best at every compute budget, and the gap generally grows with recurrence. Reading comprehension shows a different pattern: $r{=}2$ and $r{=}4$ remain competitive with the non-looped baseline, while $r{=}8$ consistently trails. Math word problems and reasoning primitives improve with compute but show no consistent ordering across recurrence counts.

To determine whether the results in Figure~\ref{fig:downstream-scaling} exceed run-to-run variation, we repeat training the compute-optimal model of each architecture at the largest budget ($C{=}2.15 {\times} 10^{19}$~FLOPs) with three seeds. Table~\ref{tab:downstream_optimum} reports the resulting means and standard deviations. The differences in parametric knowledge and reading comprehension are larger than the variation across seeds and reproduce the patterns observed across the grid. The differences on math word problems and reasoning primitives remain within run-to-run variation across training seeds.

\begin{table}[tb]
\centering
\caption{Results at the compute optimum for $C{=}2.15{\times}10^{19}$~FLOPs. The first row reports validation loss and the remaining rows report downstream continuation loss. Values are means and standard deviations across three training seeds. The final column reports $\varphi$ fitted over the full grid, with $95\%$ bootstrap intervals. ``Too noisy'' indicates that the task category does not admit a reliable fit.}
\label{tab:downstream_optimum}
\small
\setlength{\tabcolsep}{3.8pt}
\begin{tabular}{@{}lccccc@{}}
\toprule
& \multicolumn{4}{c}{Loss (nats) $\downarrow$} & \\
\cmidrule(lr){2-5}
& $r{=}1$ & $r{=}2$ & $r{=}4$ & $r{=}8$ & $\varphi$ \\
\midrule
Validation loss & $2.311 \pm 0.001$ & $2.340 \pm 0.001$ & $2.362 \pm 0.002$ & $2.403 \pm 0.003$ & $0.46$ {\footnotesize $[0.41, 0.53]$} \\
\midrule
Parametric knowledge & $2.653 \pm 0.012$ & $2.720 \pm 0.003$ & $2.754 \pm 0.009$ & $2.846 \pm 0.032$ & $0.58$ {\footnotesize $[0.50, 0.70]$} \\
Reading comprehension & $1.352 \pm 0.011$ & $1.362 \pm 0.012$ & $1.355 \pm 0.015$ & $1.394 \pm 0.016$ & $0.71$ {\footnotesize $[0.55, 0.85]$} \\
Math word problems & $1.619 \pm 0.035$ & $1.601 \pm 0.014$ & $1.605 \pm 0.013$ & $1.579 \pm 0.032$ & too noisy \\
Reasoning primitives & $0.963 \pm 0.029$ & $0.925 \pm 0.041$ & $0.940 \pm 0.051$ & $0.958 \pm 0.035$ & too noisy \\
\bottomrule
\end{tabular}
\end{table}

Finally, we test whether the recurrence equivalence exponent measured from validation loss agrees with downstream performance (Table \ref{tab:downstream_optimum}). We refit the joint scaling law using each task-category loss instead of validation loss. Parametric knowledge gives $\varphi{=}0.58$ and reading comprehension gives $\varphi{=}0.71$, compared with $\varphi{=}0.46$ for validation loss. These estimates preserve the main conclusion that recurrences provide only partial parameter equivalence, while their higher values indicate a smaller parameter-sharing cost on these downstream categories. The more challenging task categories, math word problems and reasoning primitives, do not admit reliable fits at our scale.

\subsection{Case Study Results}

We next test whether the effects of truncated BPTT and hyperconnections transfer to downstream performance. As in the baseline analysis, we obtain reliable fits only for parametric knowledge and reading comprehension. Table~\ref{tab:downstream-case-studies} compares the resulting recurrence equivalence exponents with those measured from validation loss. The downstream results preserve the ordering measured from validation loss. Truncated BPTT reduces the value of recurrence, while hyperconnections increase it. This agreement supports $\Delta\varphi$ as a diagnostic of the two case studies beyond validation loss.

\begin{table}[tb]
\centering
\caption{Recurrence equivalence exponent $\varphi$ measured from validation loss and downstream continuation loss. Truncated BPTT lowers $\varphi$ relative to the baseline, while hyperconnections raise it.}
\label{tab:downstream-case-studies}
\small
\begin{tabular}{@{}lccc@{}}
\toprule
Target & Baseline & Truncated BPTT & Hyperconnections \\
\midrule
Validation loss       & $0.46$ & $0.38$ & $0.65$ \\
Parametric knowledge  & $0.58$ & $0.50$ & $0.80$ \\
Reading comprehension & $0.71$ & $0.67$ & $0.94$ \\
\bottomrule
\end{tabular}
\end{table}

\subsection{Extrapolation Run Results}
\label{app:extrapolation}

Finally, we evaluate the $r{=}1$ and $r{=}4$ extrapolation runs from Section~\ref{sec:scaling-fits}. Both models have width $s{=}34$ and are trained on $47$B tokens, corresponding to ${\sim}4{\times}10^{20}$~FLOPs and ${\sim}20\times$ the largest budget in the baseline grid. The runs are matched by training tokens, giving the looped model a ${\sim}3\%$ training-FLOPs advantage due to its input-injection overhead. The validation-loss gap is $0.061$~nats, within the $[0.05, 0.08]$~nat range observed for $r{=}4$ across the baseline grid. Parametric knowledge again shows the largest downstream gap, while reasoning primitives remain approximately equal between the two models. Reading comprehension and math word problems show smaller gaps. Results are from a single training run, so we cannot estimate run-to-run variation at this scale.

\begin{table}[tbh]
\centering
\caption{Validation and downstream losses for the $s{=}34$ extrapolation runs trained on $47$B tokens. The gap is $r{=}4$ minus $r{=}1$, so positive values indicate that the looped model performs worse.}
\label{tab:extrapolation}
\small
\begin{tabular}{@{}lrrr@{}}
\toprule
& $r{=}1$ & $r{=}4$ & Gap \\
\midrule
Validation loss                   & $2.047$ & $2.108$ & $+0.061$ \\
\midrule
Parametric knowledge              & $2.123$ & $2.204$ & $+0.081$ \\
Reading comprehension             & $0.980$ & $1.030$ & $+0.050$ \\
Math word problems                & $1.494$ & $1.522$ & $+0.028$ \\
Reasoning primitives              & $0.802$ & $0.797$ & $-0.005$ \\
\bottomrule
\end{tabular}
\end{table}

\end{document}